\documentclass[10pt,twocolumn,letterpaper]{article}

\usepackage{cvpr}
\usepackage{times}
\usepackage{epsfig}
\usepackage{graphicx}
\usepackage{amsmath}
\usepackage{amsthm}
\usepackage{amssymb}
\usepackage{float}
\usepackage{color}
\usepackage{multirow}
\usepackage{array}
\newcolumntype{C}[1]{>{\centering\let\newline\\\arraybackslash\hspace{0pt}}m{#1}}
\usepackage{booktabs}
\usepackage{subfigure}
\usepackage{adjustbox}
\usepackage{algorithm}
\usepackage{algorithmicx}
\usepackage{algpseudocode}
\usepackage[shortlabels]{enumitem}
\theoremstyle{definition}
\newtheorem{definition}{Definition}
\usepackage[toc,page]{appendix}

\usepackage[pagebackref=true,breaklinks=true,letterpaper=true,colorlinks,bookmarks=false]{hyperref}

\cvprfinalcopy 

\def\cvprPaperID{} 

\begin{document}

\title{Symmetry and Group in Attribute-Object Compositions}

\author{Yong-Lu Li,~Yue Xu,~Xiaohan Mao,~Cewu Lu\thanks{Cewu Lu is the corresponding author, member of Qing Yuan Research Institute and MoE Key Lab of Artificial Intelligence, AI Institute, Shanghai Jiao Tong University, China.}\\
Shanghai Jiao Tong University\\
{\tt\small \{yonglu\_li, silicxuyue, mxh1999, lucewu\}@sjtu.edu.cn}
}

\maketitle
\thispagestyle{empty}

\begin{abstract}
   Attributes and objects can compose diverse compositions. To model the compositional nature of these general concepts, it is a good choice to learn them through transformations, such as coupling and decoupling. However, complex transformations need to satisfy specific principles to guarantee the rationality. In this paper, we first propose a previously ignored principle of attribute-object transformation: \textbf{Symmetry}. For example, coupling \texttt{peeled-apple} with attribute \texttt{peeled} should result in \texttt{peeled-apple}, and decoupling \texttt{peeled} from \texttt{apple} should still output \texttt{apple}. Incorporating the symmetry principle, a transformation framework inspired by group theory is built, \ie SymNet. SymNet consists of two modules, Coupling Network and Decoupling Network. With the group axioms and symmetry property as objectives, we adopt Deep Neural Networks to implement SymNet and train it in an end-to-end paradigm. Moreover, we propose a Relative Moving Distance (RMD) based recognition method to utilize the \textbf{attribute change} instead of the attribute pattern itself to classify attributes. Our symmetry learning can be utilized for the Compositional Zero-Shot Learning task and outperforms the state-of-the-art on widely-used benchmarks. Code is available at \url{https://github.com/DirtyHarryLYL/SymNet}.
\end{abstract}

\section{Introduction}
Attributes describe the properties of generic objects, \eg material, color, weight, \etc. Understanding the attributes would directly facilitate many tasks that require deep semantics, such as scene graph generation~\cite{li2017scene}, object perception~\cite{faster,maskrcnn,xu2018srda,wang2019zero,fang2019instaboost,lu2018beyond}, human-object interaction detection~\cite{hicodet,li2019hake,li2019transferable}. 
As side information, attributes can also be employed in zero-shot learning~\cite{apy,sun,awa2,cub,mit,ut}.
\begin{figure}[!ht]
	\begin{center}
		\includegraphics[width=0.45\textwidth]{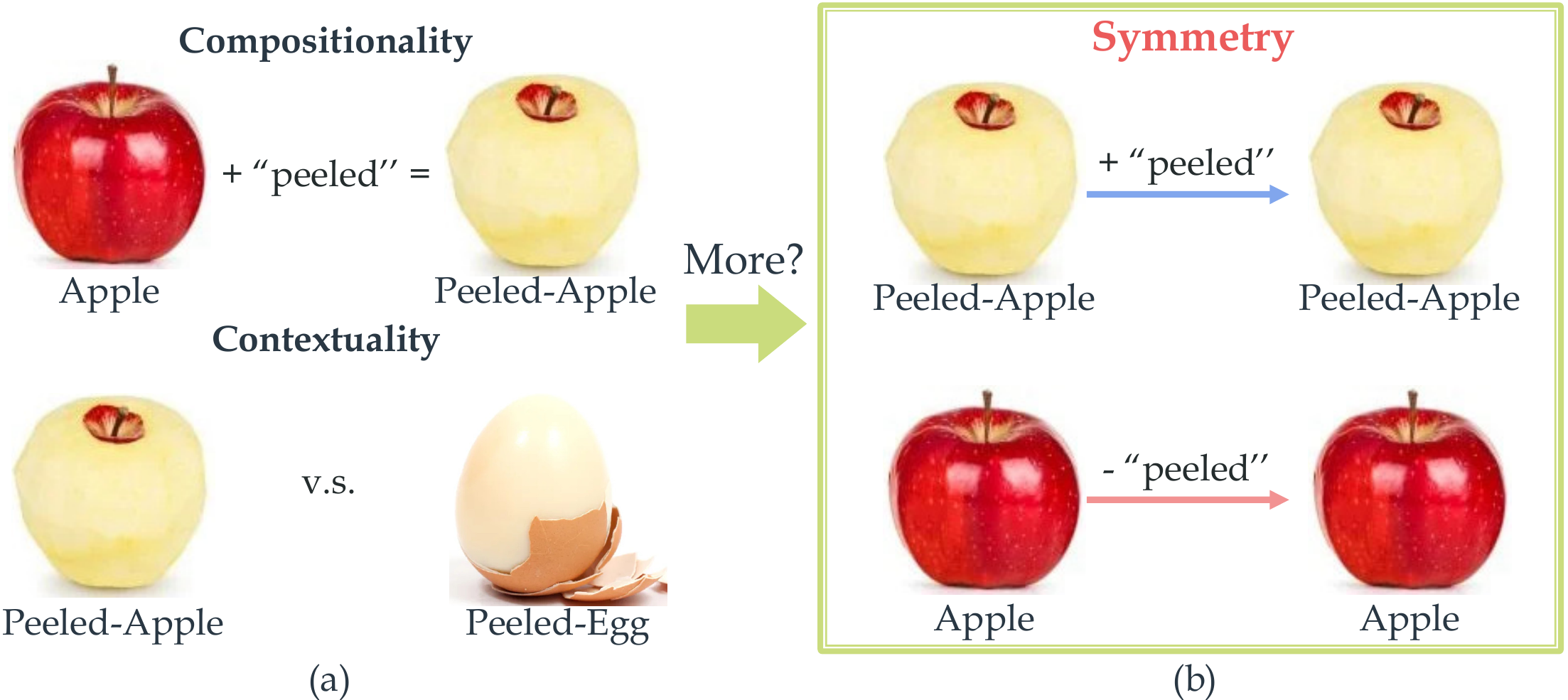}
	\end{center}
	\caption{Except for the compositionality and contextuality, attribute-object compositions also have the \textit{symmetry} property. For instance, a \texttt{peeled-apple} should not change after ``adding'' the \texttt{peeled} attribute. Similarly, an \texttt{apple} should keep the same after ``removing'' the \texttt{peeled} attribute because it does not have it.}
	\label{Figure:first-page} 
	\vspace{-0.5cm}
\end{figure}

Going along with the road of conventional classification setting, some works~\cite{awa1, relativeattr, sun, cocoattr} address attribute recognition with the typical discriminative models for objects and achieve poor performance. This is because attributes cannot be well expressed independently of the context~\cite{redwine,operator} (Fig.~\ref{Figure:first-page}(a)). 
Subsequently, researchers rethink the nature of attributes and treat them as linear operations~\cite{operator} to operate these two general concepts, \eg ``adding'' attribute to object (coupling) or ``removing'' attribute from objects (decoupling). 
Though such new insight has promoted this field, the current ``add-remove'' system is not complete and lacks an axiomatics foundation to satisfy the specific principles of nature.
In this paper, we rethink the \textit{physical} and \textit{linguistic} properties of attribute-object, and propose a previously ignored but important principle of attribute-object transformations: \textbf{symmetry}, which would promote attribute-object learning.
Symmetry depicts the invariance under transformations, \eg a circle has rotational symmetry under the rotation without changing its appearance.
The transformation that ``adding'' or ``removing'' attributes should also satisfy the symmetry:
An object should remain unchanged if we ``add'' an attribute which it already has, or ``remove'' an attribute which it does not have.
For instance, a \texttt{peeled-apple} keeps invariant if we ``add'' attribute \texttt{peeled} upon it.
Similarly, ``removing'' \texttt{peeled} from \texttt{apple} would still result in \texttt{apple}.

As shown in Fig.~\ref{Figure:first-page}(b), except the compositionality and  contextuality, the symmetry property should also be satisfied to guarantee the rationality.
In view of this, we first introduce the symmetry and propose \textbf{SymNet} to depict it.
In this work, we aim to bridge attribute-object learning and group theory. Because the elegant properties of group theory would largely help in a more principled way, given its great theoretical potential.
Thus, to cover the principles existing in transformations theoretically, the principles from group theory are borrowed to model the symmetry.
In detail, we define three transformations \{``keep'', ``add'', ``remove''\} and an operation to perform three transformations upon objects, to construct a ``group''.
To implement these, SymNet adopts Coupling Network (CoN) and Decoupling Network (DecoN) to perform coupling/adding and decoupling/removing.
On the other hand, to meet the fundamental requirements of group theory, \textit{symmetry} and the group axioms \textit{closure, associativity, identity element, invertibility element} are all implemented as the learning objectives of SymNet.
Naturally, SymNet considers the compositionality and contextuality during coupling and decoupling of various attributes and objects.
All above principles will be learned under a unified model in an end-to-end paradigm.

\begin{figure*}[!ht]
	\begin{center}
		\includegraphics[width=0.91\textwidth]{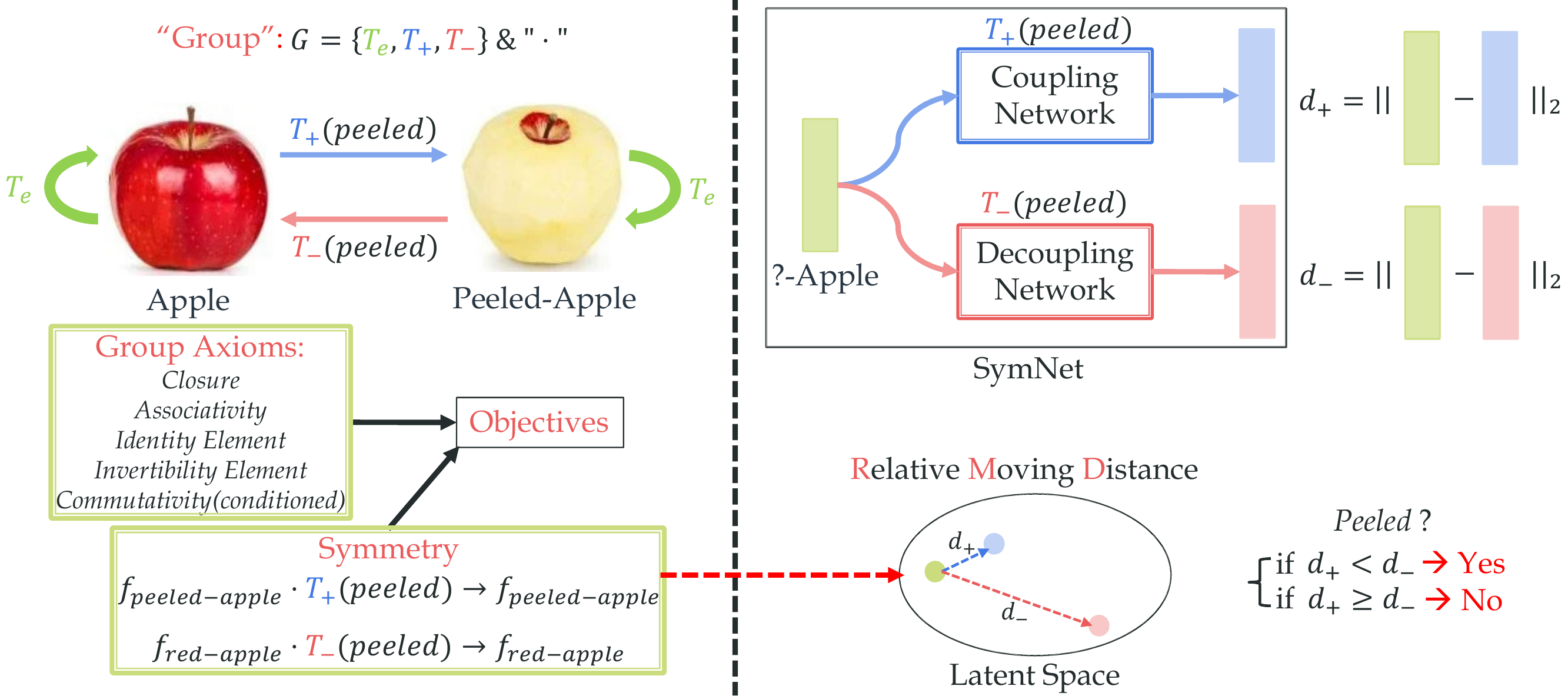}
	\end{center}
	\caption{Overview of our proposed method. We construct a ``group'' to learn the symmetry and operate the composition learning.}
	\label{Figure:overview}
	\vspace{-0.3cm}
\end{figure*}

With symmetry learning, we can apply SymNet to address the Compositional Zero-Shot Learning (CZSL), whose target is to classify the unseen compositions composed of seen attributes and objects.
We adopt a novel recognition paradigm, \textbf{R}elative \textbf{M}oving \textbf{D}istance (\textbf{RMD}) (Fig.~\ref{Figure:overview}).
That is, given a specific attribute, an object would be manipulated by the ``add'' and ``remove'' transformations parallelly in latent space. When those transformations meet the symmetry principle: if the input object already has the attribute, the output after addition should be close to the original input object, and the object after removal should be far from the input.
Contrarily, if the object does not have the given attribute, the object after removal should be closer to the input than the object after addition.
Thus, attribute classification can be accomplished concurrently by comparing the relative \textit{moving} distances between the input and two outputs.
With RMD recognition, we can utilize the robust \textit{attribute change} to classify the attributes, instead of only relying on the dramatically unstable \textit{visual attribute patterns}.
Extensive experiments show that our method achieves significant improvements on CZSL benchmarks~\cite{mit,ut}.

The main contributions of this paper are: 1) We propose a novel property of attribute-object composition transformation: symmetry, and design a framework inspired by group theory to learn it under the supervision of group axioms.
2) Based on symmetry learning, we propose a novel method to infer the attributes based on Relative Moving Distance.
3) We achieve substantial improvements in attribute-object composition zero-shot learning tasks.

\section{Related Work}
\noindent{\bf Visual Attribute.}
Visual attribute was introduced into computer vision to reduce the gap between visual patterns and object concepts, such as reducing the difficulty in object recognition~\cite{apy} or acting as an intermediate representation for zero-shot learning~\cite{awa1}. After that, attribute has been widely applied in recognition of face~\cite{celebA}, people~\cite{poselet}, pedestrian~\cite{PETA} or action~\cite{ucf101}, person Re-ID~\cite{lin2019improving, peng2017joint}, zero-shot learning~\cite{cub, sun}, caption generation~\cite{kulkarni2013babytalk, ordonez2011im2text} and so on. Therefore, attribute recognition is a fundamental problem to promote the visual concept understanding.

The typical approach for attribute recognition is to train a multi-label discriminative model same as object classification~\cite{awa1, relativeattr, sun, cocoattr}, which ignores the intrinsic properties of attributes, such as compositionality and contextuality. 
Farhadi~\etal~\cite{apy} propose a visual feature selection method to recognize the attributes, under the consideration of cross-category generalization. 
Later, some works start to consider the properties by exploiting the attribute-attribute or attribute-object correlations~\cite{hwang2011sharing, analogous, mahajan2011joint}.
Considering the contextuality of attributes, Nagarajan~\etal~\cite{operator} regard attributes as linear transformations operated upon object embeddings, and Misra~\etal~\cite{redwine} map the attributes into model weight space to attain better representations.

\noindent{\bf Compositional Zero-Shot Learning.}
CZSL is a crossing filed of compositional learning and zero-shot learning. In the CZSL setting, test compositions are unseen during training, while each component is seen in both training set and test set. Chen~\etal~\cite{analogous} construct linear classifiers for unseen compositions with tensor completion of weight vectors. 
Misra~\etal~\cite{redwine} consider that the model space is more smooth, thus project the attributes or objects into model space by training binary linear SVMs for the corresponding components. To deal with the CZSL task, it composes the attribute and object embeddings in model space as composition representation. 
Wang~\etal~\cite{tafe} address the attribute-object compositional problem via conditional embedding modification which relies on attribute word embedding~\cite{word2vec} transformation. 
Nan~\etal~\cite{genmodel} map the image features and word vectors~\cite{glove} into embedding space with the reconstruction constraint. 
Nagarajan~\etal~\cite{operator} regard attributes as linear operations for object embedding and map the image features and transformed object embeddings into a shared latent space. However, linear and explicit matrix transformation may be insufficient to represent various attribute concepts of different complexity, \eg representing ``red'' and ``broken'' as matrices with the same capacity.
Previous methods usually ignored or incompletely considered the natural principles within the coupling and decoupling of attributes and objects. In light of this, we propose a unified framework inspired by group theory to learn these important principles such as symmetry.

\section{Approach}
Fig.~\ref{Figure:overview} gives an overview of our approach. Our goal is to learn the symmetry within attribute-object compositions. Thus we can utilize it to obtain a deeper understanding of attribute-object, \eg, to address the CZSL task~\cite{mit,ut}. 
To learn the symmetry in transformations, we need a comprehensive framework to cover all principles. 
Inspired by the group theory, we define a unified model named SymNet.

We define $G = \{T_e, T_+, T_-\}$ which contains identity (``keep''), coupling (``add'') and decoupling (``remove'') transformations (Sec.~\ref{sec:define}) for each specific attribute and utilize Deep Neural Networks to implement them (Sec.~\ref{sec:implement}).
To depict symmetry theoretically, it is a natural choice to adopt group theory as the close associations between symmetry and group in physics and mathematics.
Since a group should satisfy the group axioms, \ie, \textit{closure, associativity, identity element, and invertibility element}, we construct the learning objectives based on these axioms to train the transformations (Sec.~\ref{sec:constraints}).
In addition, SymNet also satisfies the \textit{commutativity} under conditions.
With the above constraints, we can naturally guarantee compositionality and contextuality.
\textit{Symmetry} allows us to use a novel method, Relative Moving Distance, to identify whether an object has a certain attribute with the help of $T_+$ and $T_-$ (Sec.~\ref{sec:att-classification}) for CZSL task (Sec.~\ref{sec:czsl}).
 
\subsection{Group Definition}
\label{sec:define}
To depict the symmetry, we need to first define the transformations. Naturally, we need two reciprocal transformations to ``add'' and ``remove'' the attributes. 
Further, we need an axiomatic system to restrain the transformations and keep the rationality. 
Thus, we define three transformations $G = \{T_e, T_+, T_-\}$ and an operation ``$\cdot$''.
In practice, it is difficult to strictly follow the theory considering the physical and linguistic truth. 
For example, the operation between attribute transformations ``\texttt{peeled} $\cdot$ \texttt{broken}'' is odd.
Thus the ``operation'' here is defined to be operated upon object only.
\begin{definition}
\textbf{Identity transformation} $T_{e}$ keep the attributes of object.  \textbf{Coupling} transformation $T_+$ couples a specific attribute with an object. \textbf{Decoupling} transformation $T_{-}$ decouples a specific attribute from an object.
\end{definition}

\begin{definition}
Operation ``$\cdot$'' performs transformations $\{T_e,~T_+,~T_-\}$ upon object. Noticeably, operation ``$\cdot$'' is not the dot product and we use this notation to maintain the consistence with group theory. 
\end{definition}

More formally, for object $o \in \mathcal{O}$ and attribute $a^i, a^j \in \mathcal{A},~a^i\neq a^j$, where $\mathcal{O}$ denotes object set and $\mathcal{A}$ denotes attribute set, operation ``$\cdot$'' performs transformations in $G$ upon an object/image embedding:
\begin{eqnarray}
    \begin{split}
        & f_o^{i} \cdot T_+(a^j) = f_o^{ij},\\
        & f_o^{ij} \cdot T_-(a^j) = f_o^{i},\\
        & f_o^{i} \cdot T_e = f_o^{i},
    \end{split}    
\label{eq:3-T}
\end{eqnarray}
where $f_o^{i}$ means $o$ has one attribute $a^i$ and $f_o^{ij}$ means $o$ has two attributes $a^i, a^j$.
Here we do not sign a specific object category and use $o$ for simplicity. 

\begin{definition}
$G$ has the \textbf{symmetry} property if and only if $\forall a^i,~a^j \in \mathcal{A}, a^i\neq a^j$:
\begin{eqnarray}
    f_o^{i} = f_o^{i} \cdot T_+(a^i),~f_o^{i} = f_o^{i} \cdot T_-(a^j).
\label{eq:symmetry}
\end{eqnarray}
\end{definition}

\subsection{Group Implementation}
\label{sec:implement}
\begin{figure}[!t]
	\begin{center}
		\includegraphics[width=0.45\textwidth]{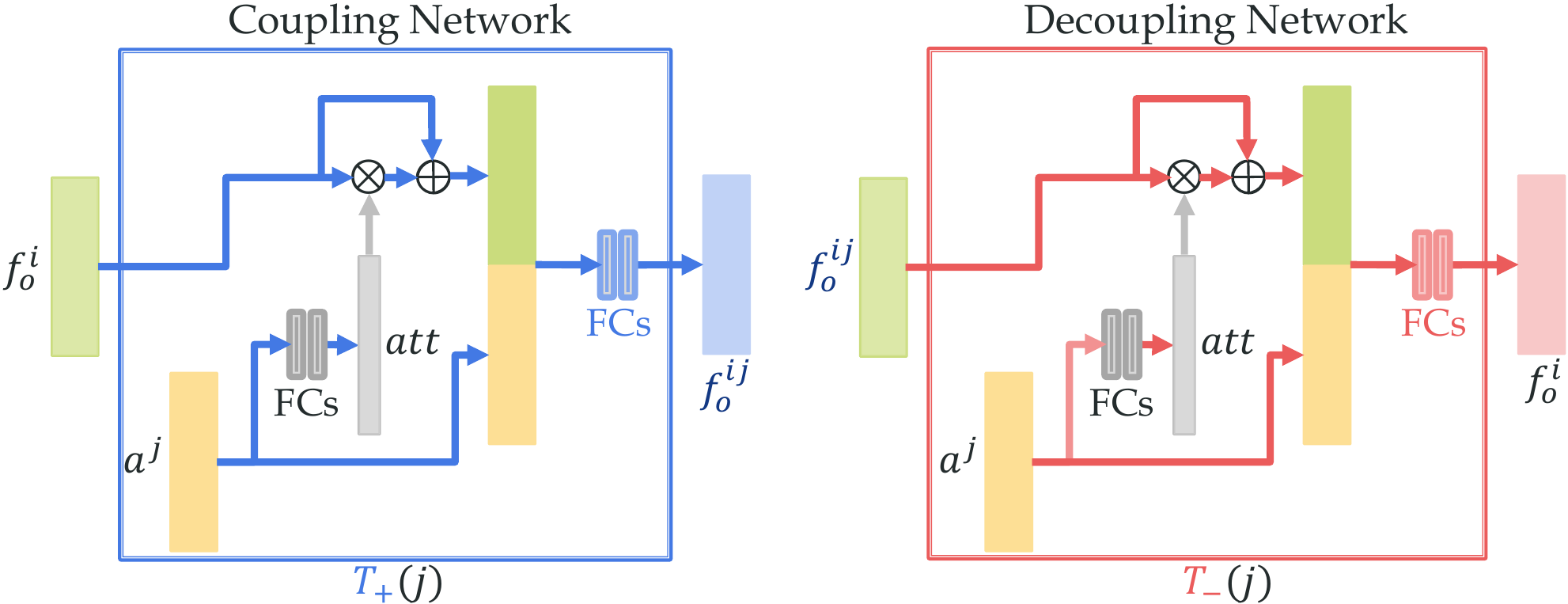}
	\end{center}
	\caption{The structure of CoN and DecoN. They take the attribute embedding to assign a specific attribute $a^j$. $f_o^i, f_o^{ij}$ are the object embeddings extracted from ResNet-18~\cite{resnet}.}
	\label{Figure:con-decon-structure} 
\vspace{-0.3cm}
\end{figure}

In practice, when performing $T_e$ upon $f_o^{i}$, we directly use $f_o^{i}$ as the $f_o^{i} \cdot T_e$ to implement the identity transformation for simplicity.
For other two transformations $T_+,~T_-$, we propose \textbf{SymNet} consists of two modules: Coupling Network (\textbf{CoN}) and Decoupling Network (\textbf{DecoN}).
CoN and DecoN have the same structure but independent weights and are trained with different tasks.
As seen in Fig.~\ref{Figure:con-decon-structure}, CoN and DecoN both take the image embedding $f_o^{i}$ of an object and the embedding of attribute $a^j$ as inputs, and output the transformed object embedding.
We use the attribute category word embeddings such as GloVe~\cite{glove} or onehot vector to represent the attributes.
$f_o^{i}$ is extracted by an ImageNet~\cite{imagenet} pre-trained ResNet~\cite{resnet} from image $I$, \ie $f_o^{i} = F_{res}(I)$.

Intuitively, attributes affect objects in different ways, \eg ``red'' changes the color, ``wooden'' changes the texture. 
In CoN and DecoN, we use an \textit{attribute-as-attention} strategy, \ie using $att=g(a^j)$ as attention, where $g(\cdot)$ means two fully-connected (FC) and a Softmax layers.
We concatenate $f_o^{i} \circ att + f_o^{i}$ with original $a^j$ as the input and use two FC layers to perform the transformation.

\subsection{Group Axioms as Objectives}
\label{sec:constraints}
According to group theory, SymNet should satisfy four group axioms: \textit{closure, associativity, identity element,} and \textit{invertibility}. Under certain conditions, attribute-object also satisfy \textit{commutativity}.
Besides, SymNet must obey the symmetry property of the attribute transformations. 

In practice, we use Deep Neural Networks to implement transformations.  
Thus, we can construct training objectives to approach the theoretic transformations following the axioms.
Considering the \textit{actual characteristics} of attribute-object compositions, we slightly adjust the axioms to construct the objectives.
Besides, there are two situations with different forms of axioms: 1) coupling or decoupling an attribute $a^i$ that the object $f_o^i$ already has, or 2) coupling or decoupling an attribute $a^j$ that object $f_o^i$ does not have. 

\noindent {\bf Symmetry.}
First of all, SymNet should satisfy the symmetry property as depicted in Eq.~\ref{eq:symmetry}, \ie, $f_o^{i} = f_o^{i} \cdot T_+(a^i), f_o^{i} = f_o^{i} \cdot T_-(a^j)$.
The symmetry is essential to keep the semantic meaning during coupling and decoupling. For example, given a \texttt{peeled-egg}, adding the attribute \texttt{peeled} again should not change the object state. Similarly, a \texttt{cup} without attribute \texttt{broken} should remain unchanged after removing \texttt{broken}.
Thus, we construct the \textbf{symmetry loss}:
\begin{eqnarray}
    \begin{split}
        \mathcal{L}_{sym} = \|f_o^{i} - f_o^{i} \cdot T_+(a^i)\|_2 +
        \|f_o^{i} - f_o^{i} \cdot T_-(a^j)\|_2.
    \end{split}
\label{eq:symmetry-loss}
\end{eqnarray}
where $a^i, a^j \in \mathcal{A}, i \neq j$. We use $L_2$ norm loss to measure the distance between two embeddings.

\noindent {\bf Closure.} 
For all elements in set $G$, their operation results should also be in $G$. In SymNet, for the attribute $a^i$ that $f_o^{i}$ has, $f_o^{i} \cdot T_+(a^i) \cdot T_-(a^i)$ should be equal to $f_o^{i} \cdot T_-(a^i)$. 
For the attribute $a^j$ that $f_o^{i}$ does not have, $f_o^{i} \cdot T_-(a^j) \cdot T_+(a^j)$ should be equal to $f_o^{i} \cdot T_+(a^j)$.
Thus, we construct:
\begin{eqnarray}
    \begin{aligned}
        \mathcal{L}_{clo} = & \|f_o^{i} \cdot T_+(a^i) \cdot T_-(a^i) -  f_o^{i} \cdot T_-(a^i)\|_2 + 
        \\ & \|f_o^{i} \cdot T_-(a^j) \cdot T_+(a^j) -  f_o^{i} \cdot T_+(a^j)\|_2,
    \end{aligned}
\label{eq:closure}
\end{eqnarray}

\noindent {\bf Identity Element.}
The properties of identity element $T_e$ are automatically satisfied since we implement $T_e$ as a skip connection, \ie $f_o^i\cdot T_*(a^i)\cdot T_e=f_o^i\cdot T_e\cdot T_*(a^i)=f_o^i\cdot T_*(a^i)$ where $T_*$ denotes any element in $G$.

\noindent {\bf Invertibility Element.}
According to the definition, $T_+$ is the invertibility element of $T_-$, vice versa. 
For the attribute $a^i$ that $f_o^{i}$ has, $f_o^{i} \cdot T_-(a^i) \cdot T_+(a^i)$ should be equal to $f_o^{i} \cdot T_e = f_o^{i}$.
For the attribute $a^j$ that $f_o^{i}$ does not have, $f_o^{i} \cdot T_+(a^j) \cdot T_-(a^j)$ should be equal to $f_o^{i} \cdot T_e = f_o^{i}$.
Therefore, we have:
\begin{eqnarray}
    \begin{aligned}
        \mathcal{L}_{inv} = & \|f_o^{i} \cdot T_+(a^j) \cdot T_-(a^j) - f_o^{i} \cdot T_e\|_2 + 
        \\ & \|f_o^{i} \cdot T_-(a^i) \cdot T_+(a^i) - f_o^{i} \cdot T_e\|_2.
    \end{aligned}
\label{eq:invertibility}
\end{eqnarray}

\noindent {\bf Associativity.} 
In view of the practical physical meaning of attribute-object compositions, we only define the operation ``$\cdot$'' that operates a transformation upon an object embedding in Sec.~\ref{sec:define}, but do not define the operation between transformations. 
Therefore, we relax the constraint here and do not construct an objective according to associativity in practice.

\noindent {\bf Commutativity.} Because of the speciality of attribute-object, SymNet satisfies the \textit{commutativity} when coupling and decoupling \textit{multiple} attributes. Thus, $f_o^{i} \cdot T_+(a^i) \cdot T_-(a^j)$ should be equal to $f_o^{i} \cdot T_-(a^j) \cdot T_+(a^i)$:
\begin{eqnarray}
    \begin{aligned}
    \mathcal{L}_{com} = \|&f_o^{i} \cdot T_+(a^i) \cdot T_-(a^j) - \\
    &f_o^{i} \cdot T_-(a^j) \cdot T_+(a^i)\|_2.
    \end{aligned}
\label{eq:commutativity}
\end{eqnarray}
Although above definitions do not strictly follow the theory, but the \textit{loosely} conducted axiom objectives still contribute to the robustness and effectiveness a lot (ablation study in Sec.~\ref{sec:czsl}) and open a door to a more theoretical way.

The last but not the least, CoN and DecoN need to keep the \textbf{semantic consistency}, \ie before and after the transformation, the \textit{object category} should not change. 
Hence we use a cross-entropy loss $\mathcal{L}_{cls}^o$ for the object recognition of the input and output embeddings of CoN and Decon.
In the same way, before and after coupling and decoupling, the \textit{attribute changes} provide the attribute classification loss $\mathcal{L}_{cls}^a$.
We use typical visual pattern-based classifiers consisting of FC layers for the object and attribute classifications.

\subsection{Relative Moving Distance} 
\label{sec:att-classification}
\begin{figure}[!t]
	\begin{center}
		\includegraphics[width=0.45\textwidth]{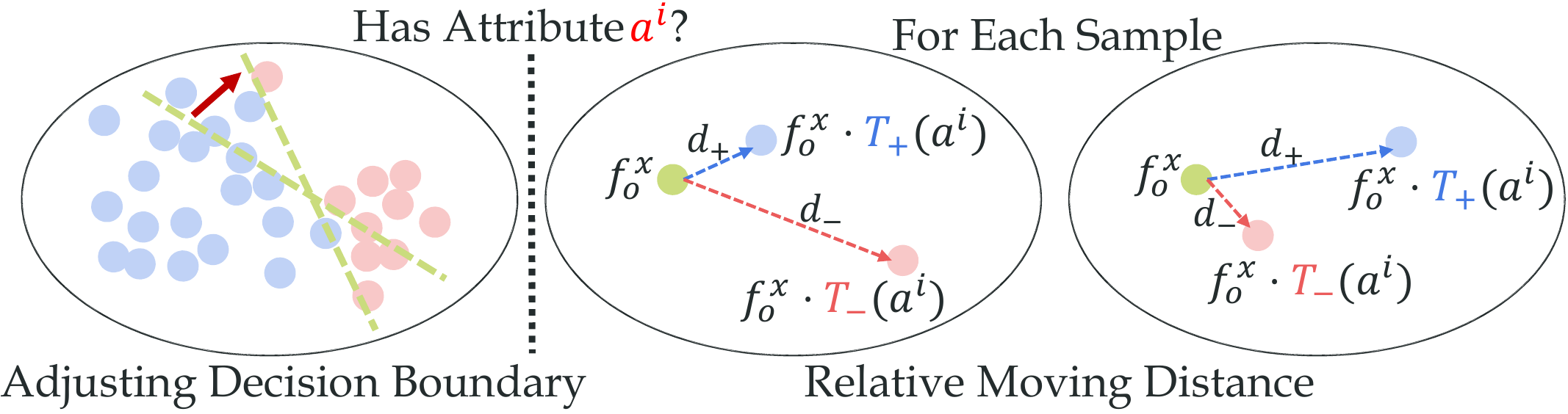}
	\end{center}
	\caption{Comparison between typical method and our Relative Moving Distance (RMD) based recognition. Previous methods mainly try to adjust the decision boundary in latent space. Our RMD based approach moves the embedding point with $T_+$ and $T_-$ and classifies by comparing their moving distances.}
	\label{Figure:rmd}
\vspace{-0.3cm}
\end{figure}

As shown in Fig.~\ref{Figure:rmd}, we utilize the Relative Moving Distance (RMD) based on the symmetry property to operate the attribute recognition.
Given an image embedding $f_o^{x}$ of an object with unknown attribute $a^x$, we first input it to both CoN and DecoN with all kinds of attributes word embeddings $\{a^1, a^2, ..., a^n\}$ where $n$ is the number of attributes.
Two transformers would take attribute embeddings as conditions and operate the coupling and decoupling in parallel, then output $2n$ transformed embeddings $\{f_o^{x} \cdot T_+(a^1), f_o^{x} \cdot T_+(a^2),..., f_o^{x} \cdot T_+(a^n)\}$ and $\{f_o^{x} \cdot T_-(a^1), f_o^{x} \cdot T_-(a^2), ..., f_o^{x} \cdot T_-(a^n)\}$.
We compute the distances between $f_o^{x}$ and the transformed embeddings:
\vspace{-3mm}
\begin{eqnarray}
    \begin{aligned}
        & d^{i}_+ = \|f_o^{x} - f_o^{x} \cdot T_+(a^i)\|_2,\\
        & d^{i}_- = \|f_o^{x} - f_o^{x} \cdot T_-(a^i)\|_2.
    \end{aligned}
\label{eq:distances}
\end{eqnarray}
To compare two distances, we define \textit{Relative Moving Distance} as $d^{i}=d^{i}_--d^{i}_+$ and perform binary classification for each attribute (Fig.~\ref{Figure:rmd}): 
1) If $d^{i} \geq 0$, \ie $f_o^{x} \cdot T_+(a^i)$ is closer to $f_o^i$ than $f_o^{x} \cdot T_-(a^i)$, we tend to believe $f_o^x$ has attribute $a^i$.
2) If $d^{i} < 0$, \ie $f_o^{x} \cdot T_-(a^i)$ is closer, we tend to predict that $f_o^x$ does not have attribute $a^i$.
Previous zero/few-shot learning methods usually classify the instances via measuring the distance between the embedded instances and \textbf{fixed} points like prototype/label/centroid embeddings.
Differently, Relative {\bf Moving} Distance compares the distance before and after applying the coupling and decoupling operation. 

\noindent \textbf{Training.}
To enhance the RMD-based classification performance, we further use a triplet loss function. Let $\mathcal{X}$ denote the set of attributes that $f_o^x$ has, the loss can be described as:
\vspace{-3mm}
\begin{eqnarray}
    \mathcal{L}_{tri} = \sum_{i}^{\mathcal{X}} [d^{i}_+ - d^{i}_- + \alpha]_{+} + \sum_{j}^{\mathcal{A}-\mathcal{X}} [d^{j}_- - d^{j}_+ + \alpha]_{+},
\label{eq:triplet-loss}
\end{eqnarray}
where $\alpha$=0.5 is triplet margin. 
$d^{i}_+$ should be less than $d^{i}_-$ for the attributes that $f_o^x$ has and greater than $d^{i}_-$ for the attributes $f_o^x$ does not have.
The total loss of SymNet is
\begin{eqnarray}
\begin{aligned}
    \mathcal{L}_{total} = &\lambda_{1}\mathcal{L}_{sym} + \lambda_{2}\mathcal{L}_{axiom} \\
    &+ \lambda_{3}\mathcal{L}_{cls}^a + \lambda_{4}\mathcal{L}_{cls}^o + \lambda_{5}\mathcal{L}_{tri},
\end{aligned}
\label{eq:total-loss}
\end{eqnarray}
where $\mathcal{L}_{axiom} = \mathcal{L}_{clo} + \mathcal{L}_{inv} + \mathcal{L}_{com}$.

\noindent \textbf{Inference.} In practice, for $n$ attribute categories, we use RMDs $d=\{d^{i}\}^n_{i=1}$ as the attribute scores, \ie $\mathcal{S}_a = \{\mathcal{S}_a^i\}^n_{i=1} = \{d^{i}\}^n_{i=1}$ and obtain attribute probability with Sigmoid function: $p_a^i = Sigmoid(\mathcal{S}_a^i)$.
Notably, we also consider the scale and use a factor $\gamma$ to adjust the score before Sigmoid.
Our method can be operated in parallel, \ie, we simultaneously compute the RMD values of $n$ attributes.
We input $[B, n, 300]$ sized tensor where $B$ is the mini-batch size and $300$ is the object embedding size. CoN and DecoN would output two $[B, n, 300]$ sized embeddings after transformation. Then we can compute RMDs $\{d^{i}\}^n_{i=1}$ at the same time.
Our method has approximately the same speed as the typical FC classifier. The inference speed from features to RMD is about 41.0 FPS and the FC classifier speed is about 45.8 FPS. The gap can be further omitted if considering the overhead of the feature extractor.

\subsection{Discussion: Composition Zero-Shot Learning}
\label{sec:czsl}
With robust and effective symmetry learning for attribute-object, we can further apply SymNet to CZSL~\cite{mit,ut}.
The goal of CZSL is to infer the unseen attribute-object pairs in test set, \ie a prediction is true positive if and only if both attribute and object classifications are accurate. The pair candidates are available during testing, thus the predictions of impossible pairs can be masked. 

We propose a novel method to address this task based on Relative Moving Distance (RMD). With Relative Moving Distance $d^{i}=d^{i}_--d^{i}_+$, the probabilities of attribute category are computed as $p_a^i = Sigmoid(d^{i})$.
For object category, we input the object embedding to 2-layer FC with Softmax to obtain the object scores $\mathcal{S}_o = \{\mathcal{S}_o^j\}^m_{j=1}$, where $m$ is the number of object categories. The object category probability $p_j=Softmax(\mathcal{S}_o^j)$ and $p_o = \{p_o^i\}^m_{j=1}$.
We then use $p_{ao}^{ij}$ to represent the probability of an attribute-object pair in test set which is composed of the $i$-th attribute category and $j$-th object.
The pair probabilities are given by $p_{ao}^{ij} = p_a^i \times p_o^j$. The impossible compositions would be masked according to the benchmarks~\cite{mit,ut}.

\section{Experiment}
\label{sec:experiment}
\subsection{Data and Metrics}
Our experiments are conducted on MIT-States~\cite{mit} and UT-Zappos50K~\cite{ut}. 
MIT-States contains 63440 images covering 245 objects and 115 attributes. Each image is attached with one single object-attribute composition label and there are 1262 possible pairs in total. We follow the setting of \cite{redwine} and use 1262 pairs/34562 images for training and 700 pairs/19191 images as the test set. 
UT-Zappos50K is a fine-grained dataset with 50025 images of shoes annotated with shoe type-material pairs. We follow the setting and split from \cite{operator}, using 83 object-attribute pairs/24898 images as train set and 33 pairs/4228 images for testing. 
The training and testing pairs are non-overlapping for both datasets, \ie the test set contains unseen attribute-object pairs composed of seen attributes and objects. 
We report the Top-1, 2, 3 accuracies on the unseen test set as evaluation metrics. 
We also evaluate our model under the generalized CZSL setting of TMN~\cite{tmn}, since the "open world" setting from \cite{operator} brings biases towards unseen pairs~\cite{chao2016empirical}.

\subsection{Baselines}
\label{sec:baseline}
We compare SymNet with baselines following \cite{redwine} and \cite{operator}, as well as previous state-of-the-arts. If not specified, the adopted methods are based on ResNet-18 backbone.

\noindent\textbf{Visual Product} trains two simple classifiers for attributes and objects independently and fuses the outputs by multiplying their margin probabilities: $P(a,o)=P(a)P(o)$. The classifiers can be either linear SVMs~\cite{redwine} or single layer softmax regression models~\cite{operator}.

\noindent\textbf{LabelEmbed (LE)} is proposed by \cite{redwine}. It combines the word vectors~\cite{glove} of attribute and object and uses 3-layer FC to transform the pair embedding into a transform matrix. The classification score is the product of transform matrix and visual feature: $\mathcal{T}\left(e_{a}, e_{b}\right)^{\top} \phi(I)$. It has three variants: 
\begin{enumerate}[nosep]
    \item \textbf{LabelEmbed Only Regression (LEOR)}~\cite{redwine} changes the target to minimize the Euclidean distance between $\mathcal{T}\left(e_{a}, e_{b}\right)$ and the weight of pair SVM classifier $w_{ab}$.
    \item \textbf{LabelEmbed With Regression (LE+R)}~\cite{redwine} combines the losses of LE and LEOR aforementioned.
    \item \textbf{LabelEmbed+}~\cite{operator} embeds the attribute, object vectors, and image features into a semantic space and also optimizes the input representations during training.
\end{enumerate}

\noindent\textbf{AnalogousAttr}~\cite{analogous} trains linear classifiers for seen compositions and uses tensor completion to generalize to the unseen pairs. We report the reproduced results from \cite{operator}.

\noindent\textbf{Red Wine}~\cite{redwine} uses SVM weights as the attribute or object embeddings to replace the word vectors in LabelEmbed.

\noindent\textbf{AttrOperator}~\cite{operator} regards attributes as linear transformations and object word vectors~\cite{glove} after transformation as pair embeddings. It takes the pair with the closest distance to the image feature as the recognition result. Besides the top-1 accuracy directly reported in \cite{operator}, we evaluate the top-2, 3 accuracies with the open-sourced code.

\noindent\textbf{TAFE-Net}~\cite{tafe} uses word vectors~\cite{word2vec} of attribute-object pair as task embedding of its meta learner. It generates a binary classifier for each existing composition. We report the results based on VGG-16 which is \textit{better} and more complete than the result based on ResNet-18.

\noindent\textbf{GenModel}~\cite{genmodel} projects the visual features of images and semantic language embeddings of pairs into a shared latent space. The prediction is given by comparing the distance between visual features and all candidate pair embeddings.

\noindent\textbf{TMN}~\cite{tmn} adopts a set of small FC-based modules and configure them via a gating function in a task-driven way. It can be generalized to unseen pairs via re-weighting these primitive modules.

\subsection{Implementation Details}
\begin{table}[t]
	\centering
	\small
	\adjustbox{width=\linewidth}{
		\begin{tabular}{lccc|ccc}
			\toprule
			\multirow{2}{*}{Method}&  \multicolumn{3}{c}{MIT-States}& \multicolumn{3}{c}{UT-Zappos}\\
			& Top-1 & Top-2 & Top-3 & Top-1 & Top-2 & Top-3 \\ 
			\midrule
			Visual Product~\cite{redwine}  & 9.8/13.9$^*$ & 16.1 & 20.6 & 49.9$^*$ & / & / \\
			LabelEmbed (LE)~\cite{redwine} & 11.2/13.4$^*$& 17.6 & 22.4 & 25.8$^*$ & / & / \\
			~- LEOR~\cite{redwine}            & 4.5          & 6.2  & 11.8 &  /       & / & / \\
			~- LE + R~\cite{redwine}          & 9.3          & 16.3 & 20.8 &  /       & / & / \\
			~- LabelEmbed+~\cite{operator}    & 14.8*         &  /   &  /   & 37.4*& / & / \\
			\midrule
			AnalogousAttr~\cite{analogous} & 1.4          &  /   &  /   & 18.3  &  /  &  /  \\
			Red Wine~\cite{redwine}        & 13.1         & 21.2 & 27.6 & 40.3  &  /  &  /  \\ 
			AttOperator~\cite{operator}    & 14.2         & 19.6 & 25.1 & 46.2  & 56.6 & 69.2 \\
			TAFE-Net~\cite{tafe}           & 16.4         & 26.4 & 33.0 & 33.2  &  /  &  /  \\
			GenModel~\cite{genmodel}       & 17.8         &  /   &  /   & 48.3  &  /  &  /  \\
			\midrule
			SymNet (Ours) & \textbf{19.9} & \textbf{28.2} & \textbf{33.8} & \textbf{52.1}  &\textbf{67.8} &  \textbf{76.0}\\ 
			\bottomrule
	\end{tabular}}
	\caption{\small Results of CZSL on MIT-States and UT-Zappos.} 
	\label{tab:mit-ut}
	\vspace{-0.5cm}
\end{table}
For two datasets, we use ImageNet pre-trained ResNet-18~\cite{resnet} as the backbone to extract image features and do not fine-tune it following previous methods. We use the 300-dimensional pre-trained GloVe~\cite{glove} vectors as the word embeddings.
The 512-dimensional ResNet-18 feature is first transformed to 300-dimensional by a single FC. The main modules of our SymNet, CoN and DecoN, have the same structures but independent weights as depicted in Fig.~\ref{Figure:con-decon-structure}: two FC layers of sizes 768/300 with Sigmoid convert the attribute embedding to 300-dimensional attention and be multiplied to the input image representation. The representation after attention is concatenated to the attribute embedding and then compressed to the original dimension by the other two 300-sized FC layers. Each hidden FC in CoN and DecoN is followed by BatchNorm and ReLU layers.

For each training image, we randomly sample another image with the same object label but different attribute as the negative sample to compute the losses (Sec.~\ref{sec:constraints}).
We train SymNet with SGD optimizer on single NVIDIA GPU. 
We use cross-validation to determine the hyper-parameters, \eg, learning rate, weights, epochs.
For MIT-States, the model is trained with learning rate 5e-4 and batch size 512 for 320 epochs. 
The loss weights are $\lambda_1=0.05,\lambda_2=0.01,\lambda_3=1,\lambda_4=0.01, \lambda_5=0.03$. 
For UT-Zappos, the model is trained with learning rate 1e-4 and batch size 256 for 600 epochs. The loss weights are $\lambda_1=0.01,\lambda_2=0.03,\lambda_3=1,\lambda_4=0.5, \lambda_5=0.5$.
Notably, the weights on two datasets are different. Because MIT-States has diverse attributes and objects, while UT-Zappos contains similar fine-grained shoes. Different range and scale lead to distinct embedding spaces and different parameters for RMD.

\subsection{Compositional Zero-Shot Learning}
To evaluate the symmetry learning in compositional zero-shot task, we conduct experiments on widely-used benchmarks: MIT-States~\cite{mit} and UT-Zappos~\cite{ut}.

\noindent{\bf Composition Learning.}
The results of CZSL are shown in Tab.~\ref{tab:mit-ut}, where the first five rows are baselines from \cite{redwine, operator} (the scores with $*$ are reproduced by \cite{operator}, the others are from \cite{redwine}). 
SymNet outperforms all baselines on two benchmarks. 
Although we use a simple product to compose the attribute and object scores, we still achieve 2.1\% and 3.8\% improvements over the state-of-the-art~\cite{genmodel} on two benchmarks respectively.
On UT-Zappos, most previous approaches do not surpass the Visual Product baseline, while ours outperforms it by 2.2\%.
To further evaluate our SymNet, we additionally conduct the comparison on generalized CZSL setting from recent state-of-the-art TMN~\cite{tmn}.
The results are shown in Tab.~\ref{tab:gczsl-result}. SymNet also outperforms previous methods significantly, which strongly proves the effectiveness of our method. 

\begin{table}[t]
	\centering
	\small
	\resizebox{0.45\textwidth}{!}{
		\begin{tabular}{l|ccc|ccc|ccc}
			\toprule
			\multirow{2}{*}{Model} & \multicolumn{3}{|c|}{Val AUC}& \multicolumn{3}{|c|}{Test AUC} & \multirow{2}{*}{Seen} & \multirow{2}{*}{Unseen} & \multirow{2}{*}{HM} \\
			& 1 & 2 & 3 & 1 & 2 & 3 & & & \\
			\midrule
			AttOperator~\cite{operator}  & 2.5 & 6.2 & 10.1 & 1.6 & 4.7 & 7.6 & 14.3    & 17.4 & 9.9  \\
			Red Wine~\cite{redwine}      & 2.9 & 7.3 & 11.8 & 2.4 & 5.7 & 9.3 & 20.7    & 17.9 & 11.6 \\
			LabelEmbed+~\cite{operator}  & 3.0 & 7.6 & 12.2 & 2.0 & 5.6 & 9.4 & 15.0    & 20.1 & 10.7 \\
			GenModel~\cite{genmodel}     & 3.1 & 6.9 & 10.5 & 2.3 & 5.7 & 8.8 & \textbf{24.8}    & 13.4 & 11.2 \\
			TMN~\cite{tmn}               & 3.5 & 8.1 & 12.4 & 2.9 & 7.1 & 11.5& 20.2    & 20.1 & 13.0 \\
			\midrule
			SymNet (Ours) & \textbf{4.3} & \textbf{9.8} & \textbf{14.8} & \textbf{3.0} & \textbf{7.6} & \textbf{12.3} & 24.4 & \textbf{25.2} & \textbf{16.1} \\
			\bottomrule
	\end{tabular}}
	\caption{\small Results of generalized CZSL on MIT-States. All methods (Sec.~\ref{sec:baseline}) use ResNet-18~\cite{resnet} as the backbone.}
	\label{tab:gczsl-result}
\end{table}

\begin{table}[!t]
	\begin{center}
		\resizebox{0.4\textwidth}{!}{
		\begin{tabular}{lccccc}
		\hline  
		&  \multicolumn{2}{c}{MIT-States} & \multicolumn{2}{c}{UT-Zappos} \\
	    Method & Attribute & Object & Attribute & Object\\
	    \hline 
	    AttrOperator~\cite{operator} & 14.6 & 20.5 & 29.7 & 67.5 \\
	    GenModel~\cite{genmodel} & 15.1 & 27.7 & 18.4 & 68.1 \\
	    \hline
	    SymNet  &  \textbf{18.9} & 28.8  & \textbf{38.0} & 65.4 \\
	    \hline
		\end{tabular}}
	\end{center}
	\caption{Attribute learning results on two benchmarks.}
	\label{tab:att}
\vspace{-0.5cm}
\end{table}

\noindent{\bf Attribute Learning.}
We also compare the attribute accuracy alone on two benchmarks in Tab.~\ref{tab:att}.
We reproduce the results of AttrOperator~\cite{operator} with its open-sourced code. For all methods involved, the individual attribute and object accuracy do not consider the relations between attributes and objects. 
The object recognition module of our method is a simple 3-layer MLP classifier with the visual image features from ResNet-18 backbone.
SymNet outperforms previous methods by a large margin, \ie 3.8\% on MIT-States and 8.3\% on UT-Zappos. 
Our RMD-based attribute recognition is particularly effective.
In addition, our object classification performance is comparable to AttrOperator~\cite{operator} and GenModel~\cite{genmodel}. 
Accordingly, the main contribution of the CZSL improvement of SymNet comes from attribute learning rather than object recognition.

\subsection{Image Retrieval after Attribute Manipulation}
\begin{figure}[!ht] 
	\begin{center} 
		\includegraphics[width=0.45\textwidth]{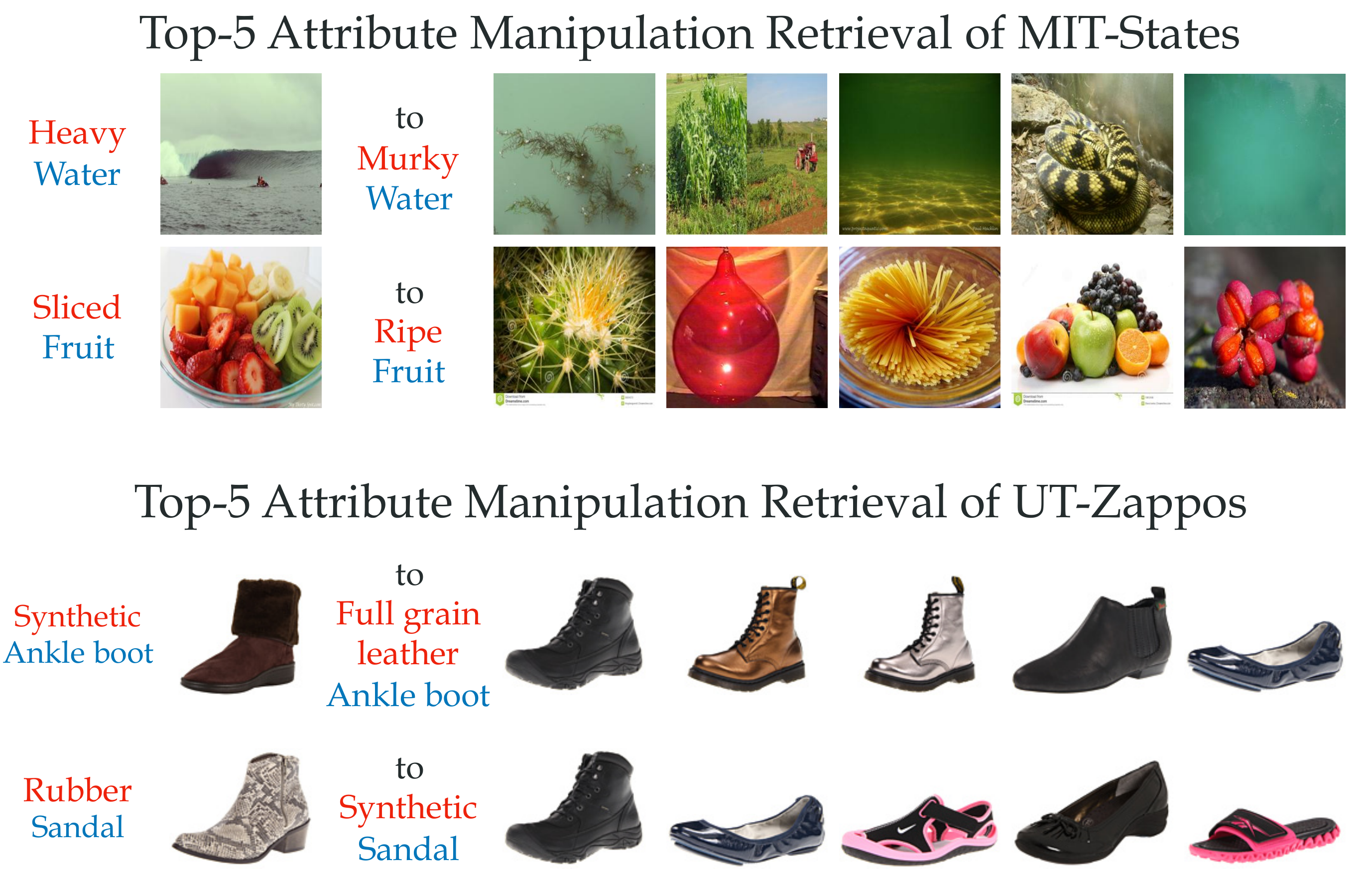}
	\end{center}
	\caption{Image Retrieval on MIT-States, UT-Zappos. We conduct the retrieval after the \textit{attribute manipulation}.}
	\label{Figure:retrieval} 
\vspace{-0.5cm}
\end{figure}
To qualitatively evaluate SymNet, we further report the image retrieval results after attribute manipulation. 
We first train SymNet on MIT-States or UT-Zappos, then use trained CoN and DeCoN to manipulate the image embeddings. 
For an image with pair label $(a,o)$, we remove the attribute $a$ with DeCoN and add an attribute $b$ with CoN, then we retrieve the top-5 nearest neighbors of the manipulated embeddings. 
This task is much more difficult than the normal attribute-object retrieval~\cite{operator,redwine,tmn} because of the complex semantic manipulation and recognition.
Retrieval results are shown in Fig.~\ref{Figure:retrieval}, where the imaged on the left are original ones and right are the nearest neighbors after manipulation.
SymNet is capable of retrieving a certain number of correct samples among top-5 nearest neighbors, especially in a fine-grained dataset like UT-Zappos. 
This suggests that our model has well exploited the learned symmetry in attribute transformation and learned the contextuality and compositionality of attributes. 

\begin{figure*}[!ht]
    \begin{center}
        \begin{minipage}{.49\linewidth}
            \centerline{\includegraphics[width=\linewidth]{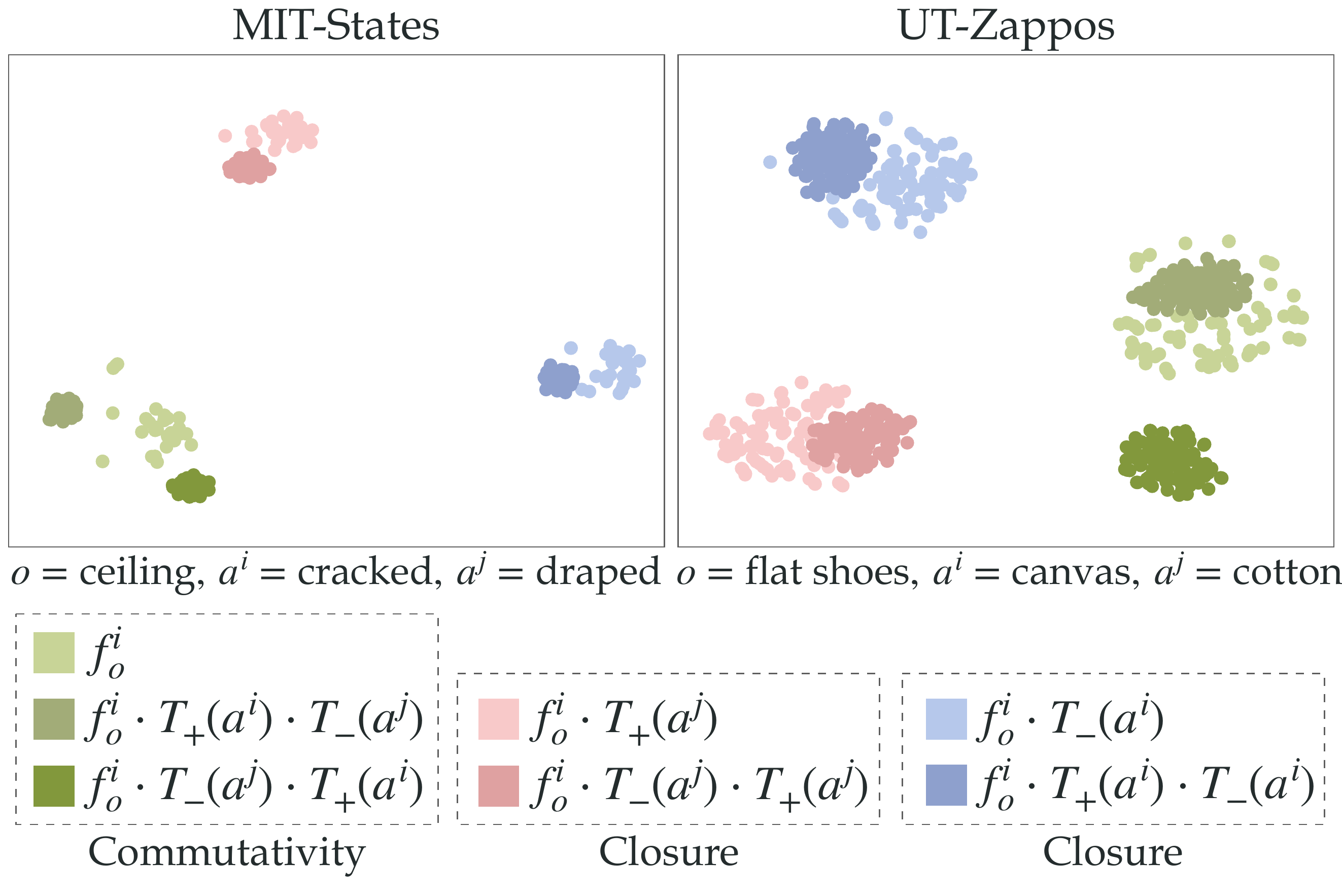}}
            \centerline{\small (a) Closure and Commutativity}
        \end{minipage}
        \hfill
        \begin{minipage}{.49\linewidth}
            \centerline{\includegraphics[width=\linewidth]{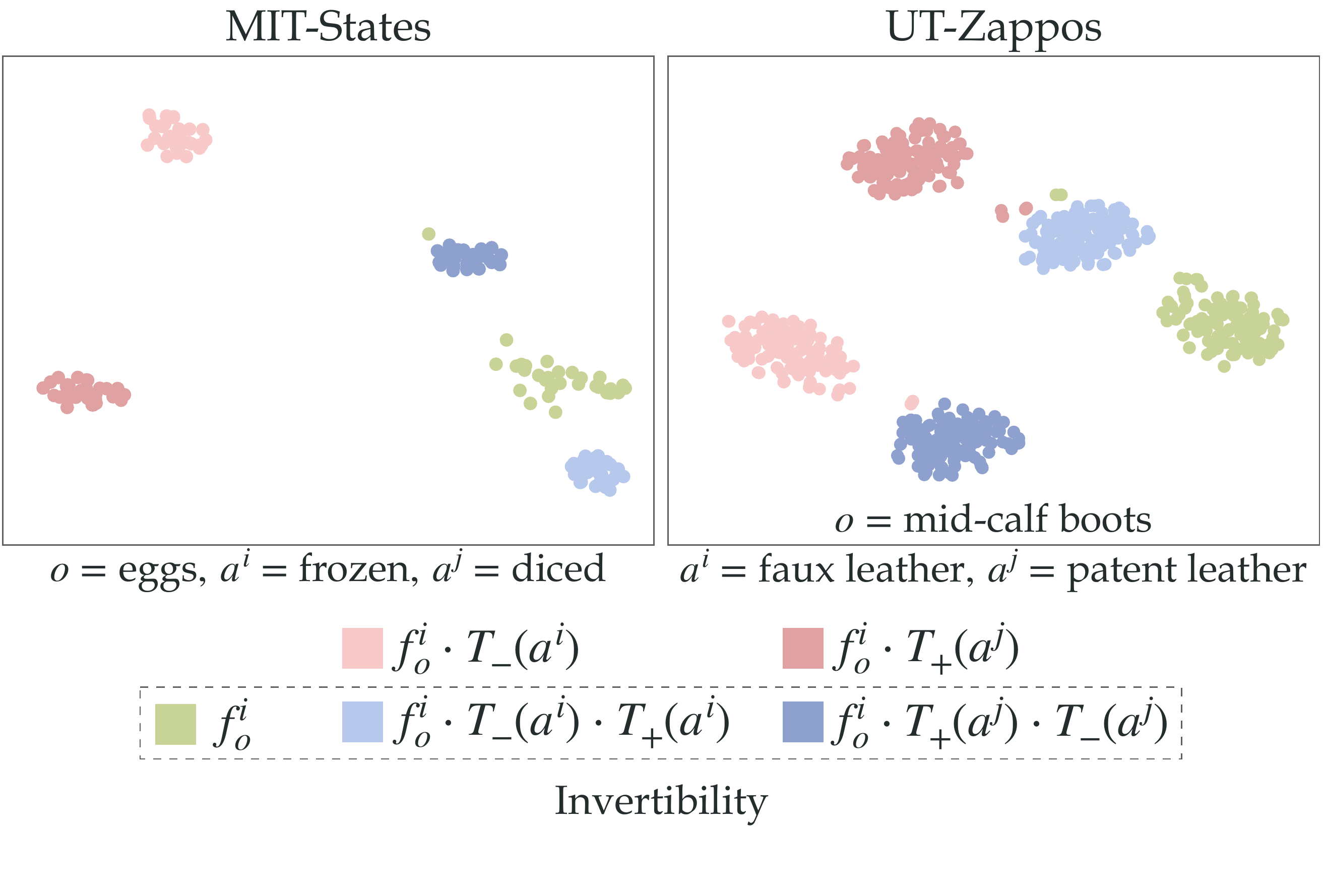}}    
            \centerline{\small (b) Invertibility}
        \end{minipage}
        \vfill
        \begin{minipage}{.49\linewidth}
            \centerline{\includegraphics[width=\linewidth]{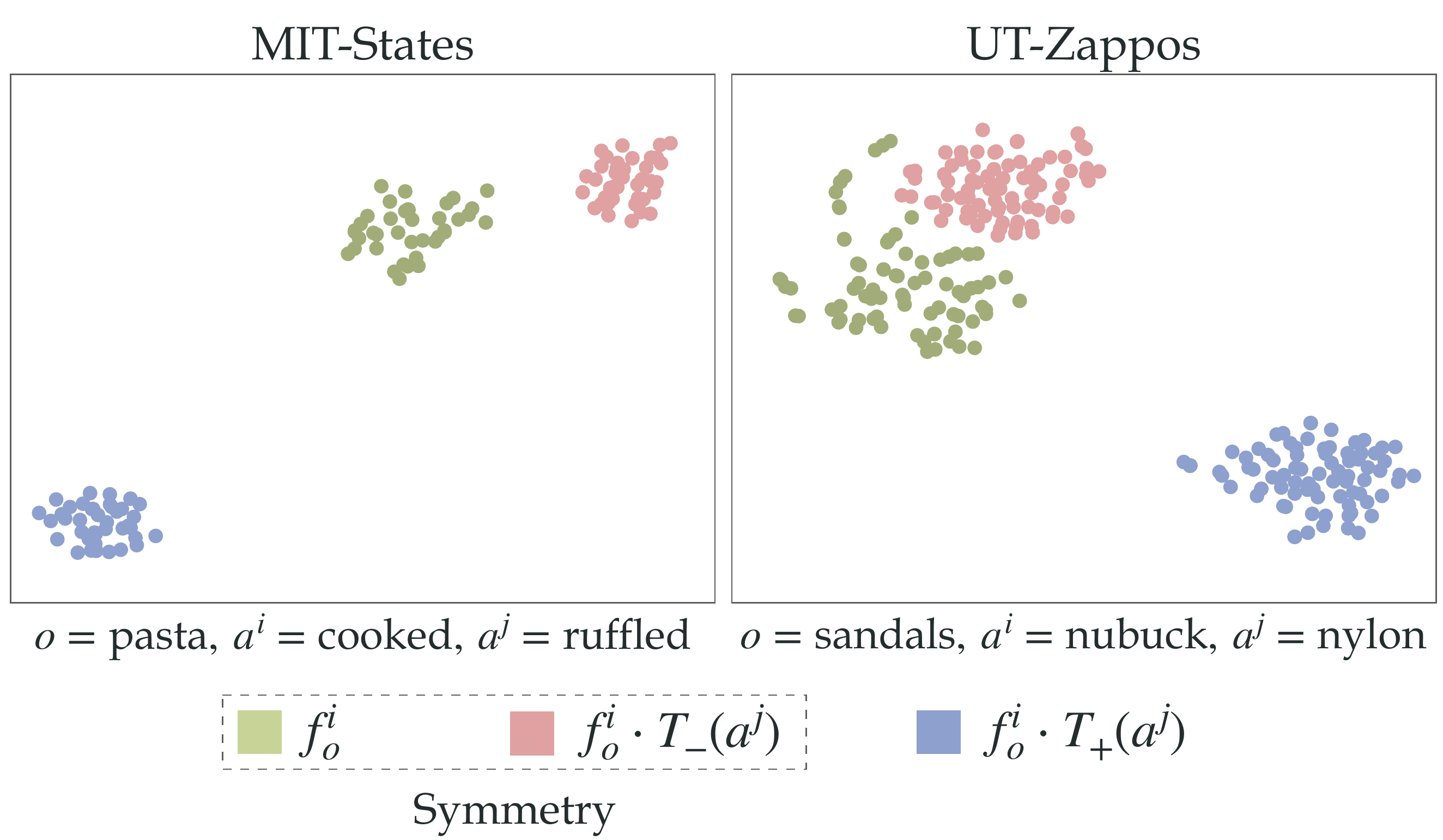}}
            \centerline{\small (c) Symmetry-1}
        \end{minipage}
        \hfill
        \begin{minipage}{.49\linewidth}
            \centerline{\includegraphics[width=\linewidth]{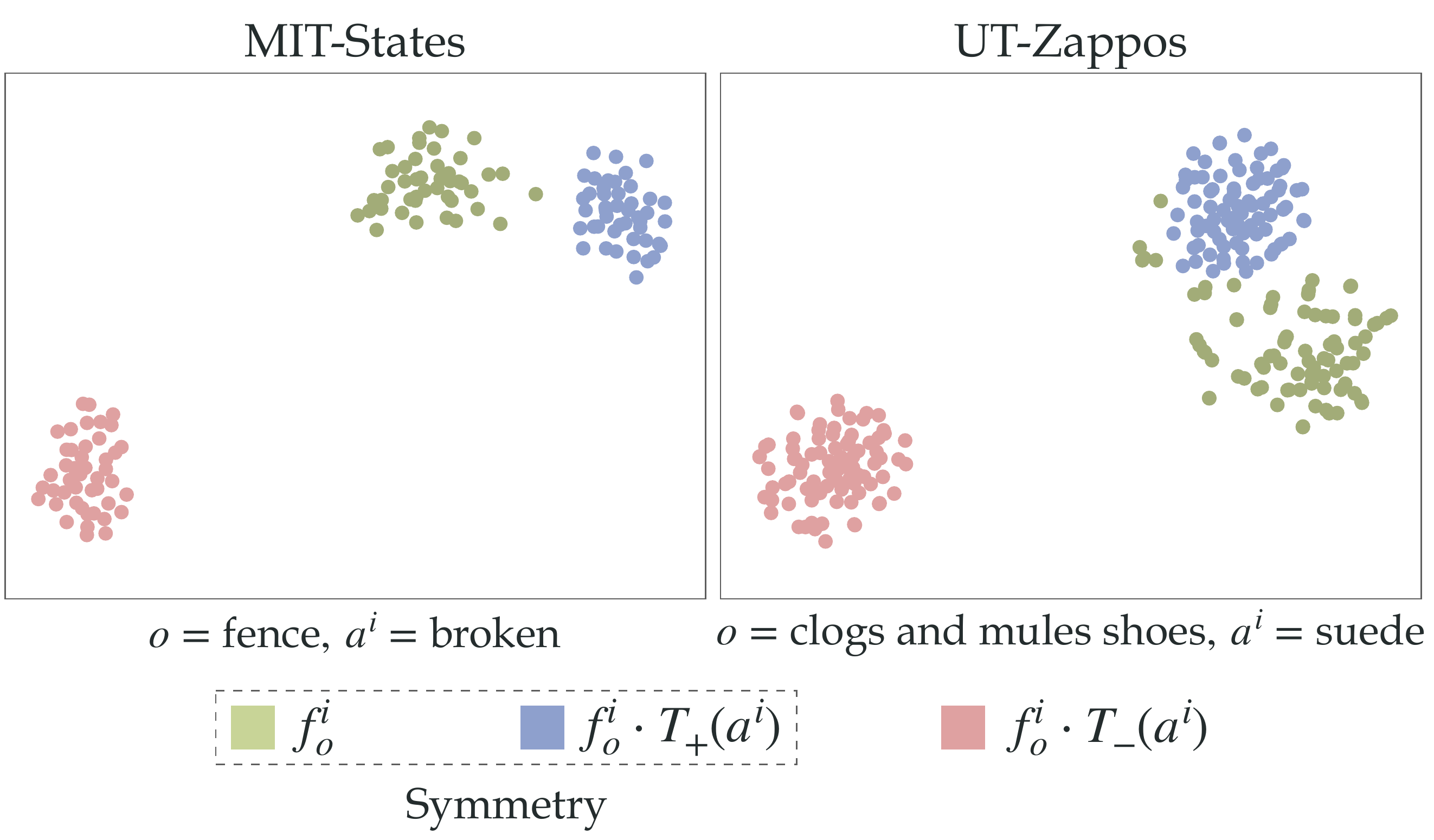}}
            \centerline{\small (d) Symmetry-2}
        \end{minipage}
	\end{center}
	\caption{Visualization of symmetry and the group axioms by t-SNE~\cite{tsne}. The points with colors in a same dotted box should be close.} 
	\label{Figure:tsne} 
\vspace{-0.4cm}
\end{figure*}

\subsection{Visualization in Latent Space}
To verify the robustness and principles in transformations, we use t-SNE~\cite{tsne} to visualize the image embeddings before or after transformations in latent space in Fig.~\ref{Figure:tsne}.
Specifically, we first visualize the group axioms related transformations:
1) \textbf{Closure} is verified by comparing \{$f_o^{i} \cdot T_+(a^i) \cdot T_-(a^i)$ v.s. $f_o^{i} \cdot T_-(a^i)$\} and \{$f_o^{i} \cdot T_-(a^j) \cdot T_+(a^j)$ v.s.  $f_o^{i} \cdot T_+(a^j)$\}.
2) \textbf{Invertibility} is verified by comparing \{$f_o^{i} \cdot T_+(a^j) \cdot T_-(a^j)$ v.s. $f_o^{i} \cdot T_e$\} and \{$f_o^{i} \cdot T_-(a^i) \cdot T_+(a^i)$ v.s. $f_o^{i} \cdot T_e$\}.
3) \textbf{Commutativity} is verified by comparing \{$f_o^{i} \cdot T_+(a^i) \cdot T_-(a^j)$ v.s. $f_o^{i} \cdot T_-(a^j) \cdot T_+(a^i)$\}. The results are shown in Fig.~\ref{Figure:tsne} (a,b). 
We observe that SymNet can robustly operate the transformations and the axiom objectives are well satisfied during embedding transformations.

Then, to verify the \textbf{symmetry} property, we visualize the sample embeddings in Relative Moving Space in Fig.~\ref{Figure:tsne}(c,d):
1) for the sample $f_o^i$ which do not have attribute $a^j$, $f_o^i \cdot T_+(a^j)$ should be far from $f_o^i$. On the contrary, $f_o^i \cdot T_-(a^j)$ are relatively close to $f_o^i$ because of the symmetry.
2) For the sample $f_o^i$ with attribute $a^i$, $f_o^i \cdot T_+(a^i)$  should be close to $f_o^i$ and $f_o^i \cdot T_-(a^i)$ should be far from $f_o^i$.
We can also find that the relative moving distance rules are all satisfied, \ie the symmetry is well learned by our SymNet.

\begin{table}[t]
	\centering
	\small
	\adjustbox{width=\linewidth}{
		\begin{tabular}{lccc|ccc}
			\toprule
			\multirow{2}{*}{Method} &  \multicolumn{3}{c}{MIT-States}& \multicolumn{3}{c}{UT-Zappos}\\
			& Top-1 & Top-2 & Top-3 & Top-1 & Top-2 & Top-3 \\
			\midrule
			SymNet & \textbf{19.9} & \textbf{28.2} & \textbf{33.8} & \textbf{52.1}  &\textbf{67.8} &  \textbf{76.0}\\ 
			\midrule
			SymNet w/o $\mathcal{L}_{sym}$   & 18.3 & 27.5 & 33.4 & 51.1 & 67.0 & 76.0 \\
			SymNet w/o $\mathcal{L}_{axiom}$ & 16.9 & 25.5 & 30.9 & 47.6 & 65.4 & 73.6 \\
			SymNet w/o $\mathcal{L}_{inv}$   & 17.9 & 26.7 & 32.5 & 50.8 & 67.4 & 76.1 \\
			SymNet w/o $\mathcal{L}_{com}$   & 17.8 & 27.0 & 32.7 & 51.2 & 67.6 & 75.8 \\
			SymNet w/o $\mathcal{L}_{clo}$   & 18.0 & 27.0 & 32.8 & 51.1 & 67.2 & 76.0 \\
			SymNet w/o $\mathcal{L}_{cls}$   & 10.3 & 18.9 & 25.9 & 28.7 & 51.2 & 65.2 \\ 
			SymNet w/o $\mathcal{L}_{tri}$   & 17.8 & 26.8 & 32.6 & 49.2 & 65.3 & 74.2 \\
			SymNet w/o $\mathcal{L}_{sym}$ \& $\mathcal{L}_{tri}$ & 17.7 & 27.0 & 33.0 & 50.1 & 66.1 & 75.6 \\
			SymNet w/o $\mathcal{L}_{tri}$ \& $\mathcal{L}_{cls}$ & 10.5 & 19.4 & 26.7 & 28.6 & 51.4 & 65.6 \\
			SymNet w/o $\mathcal{L}_{sym}$ \& $\mathcal{L}_{cls}$ & 9.3 & 17.0 & 22.7 & 27.4 & 48.2 & 64.1 \\
			SymNet only $\mathcal{L}_{sym}$ & 9.4 & 16.9 & 22.5 & 20.4 & 38.9 & 53.5 \\
			SymNet w/o attention & 18.0 & 26.9 & 32.7 & 48.5 & 65.0 & 75.6 \\
			\midrule
			SymNet $L_1$ dist. & 7.1 & 11.2 & 14.3 & 37.5 & 53.3 & 62.3 \\
			SymNet $Cos$ dist. & 11.3 & 20.7 & 28.5 & 18.7 & 41.1 & 60.0 \\
			\bottomrule
	\end{tabular}}
	\caption{\small Results of ablation studies.}
	\label{tab:abl}
    \vspace{-0.5cm}
\end{table}

\subsection{Ablation Study}
To evaluate different components of our method, we design ablation studies and report the results in Tab.~\ref{tab:abl}.

\noindent\textbf{Objectives}.
To evaluate the objectives constructed from group axioms and the core principle symmetry, we conduct tests of these objectives by removing them.
In Tab.~\ref{tab:abl}, SymNet shows obvious degradations without the constraints of these principles.
This is in line with our assumption that a transformation framework that covers the essential principles can largely promote compositional learning.

\noindent\textbf{Attention}.
Removing the attention module drops 1.9\% and 3.6\% accuracy on two benchmarks.

\noindent\textbf{Distance Metrics}.
SymNet with other distance metrics, \ie, $L_1$ and cosine distances, perform much worse than $L_2$.

\section{Conclusion}
In this paper, we propose the symmetry property of attribute-object compositions.
Symmetry reveals profound principles in composition transformations.
To an object, giving it an attribute it already has, or erasing an attribute it does not have, would all result in the same object.
To learn the symmetry, we construct a framework inspired by group theory to couple and decouple attribute-object compositions, and use group axioms and symmetry as the learning objectives.
When applied to CZSL, our method achieves state-of-the-art performance.
In the future, we consider to study the transformation with varying degrees, \eg, \texttt{not-peeled}, \texttt{half-peeled} and \texttt{totally-peeled} and apply SymNet to GAN-related tasks.

{\small
\paragraph{Acknowledgement:} This work is supported in part by the National Key R\&D Program of China, No. 2017YFA0700800, National Natural Science Foundation of China under Grants 61772332 and Shanghai Qi Zhi Institute. 
}

{\small
\bibliographystyle{ieee_fullname}
\bibliography{egbib}
}

\clearpage
\onecolumn
\begin{appendices}

\section{Image Retrievals}
\begin{figure*}[!ht]
	\begin{center}
        \begin{minipage}{.33\linewidth}
            \centerline{\includegraphics[width=\linewidth]{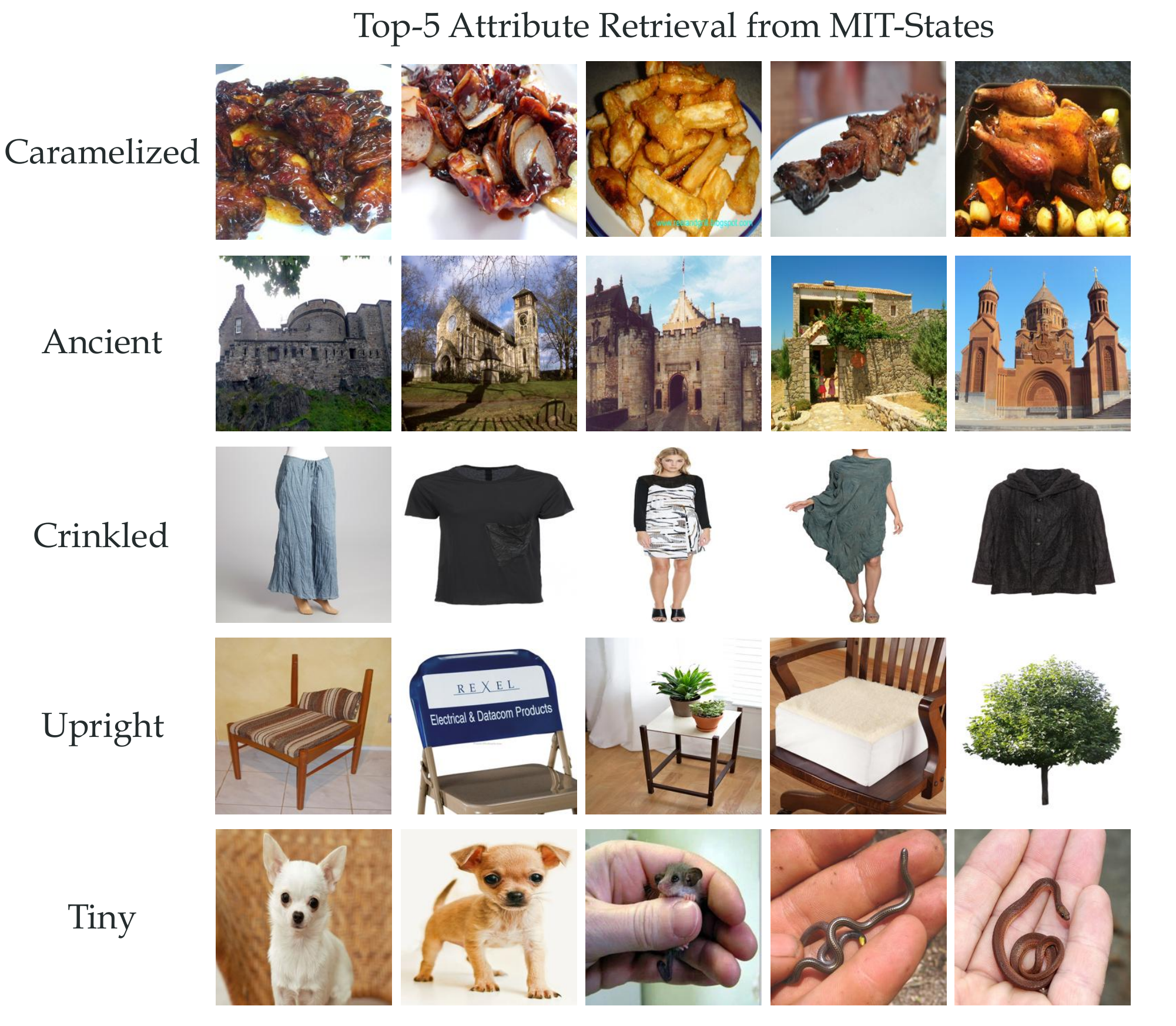}}
            \centerline{(a-1) MIT-States, attributes}
        \end{minipage}
        \hfill
        \begin{minipage}{.33\linewidth}
            \centerline{\includegraphics[width=\linewidth]{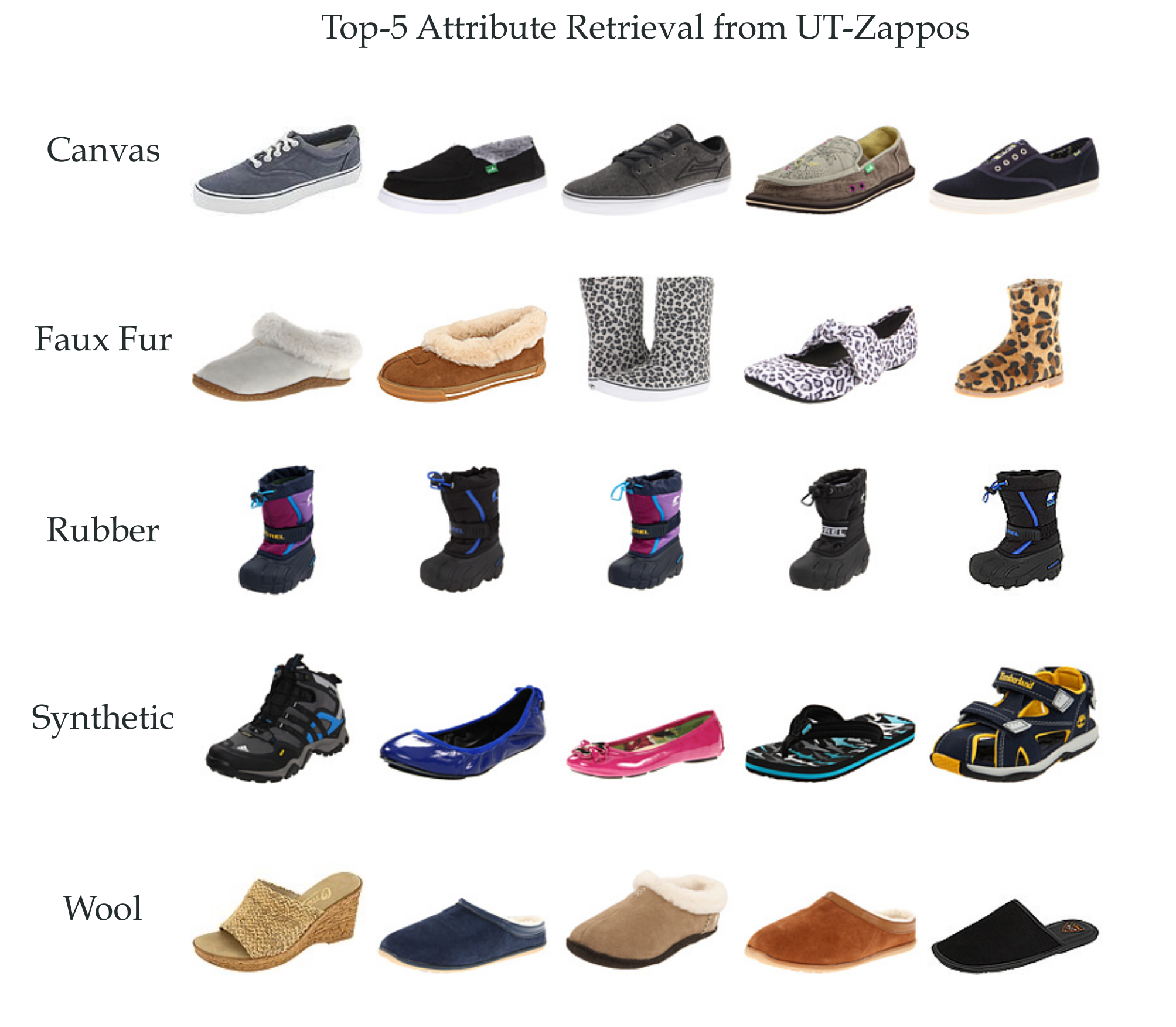}}
            \centerline{(b-1) UT-Zappos, attributes}
        \end{minipage}
        \hfill
        \begin{minipage}{.33\linewidth}
            \centerline{\includegraphics[width=\linewidth]{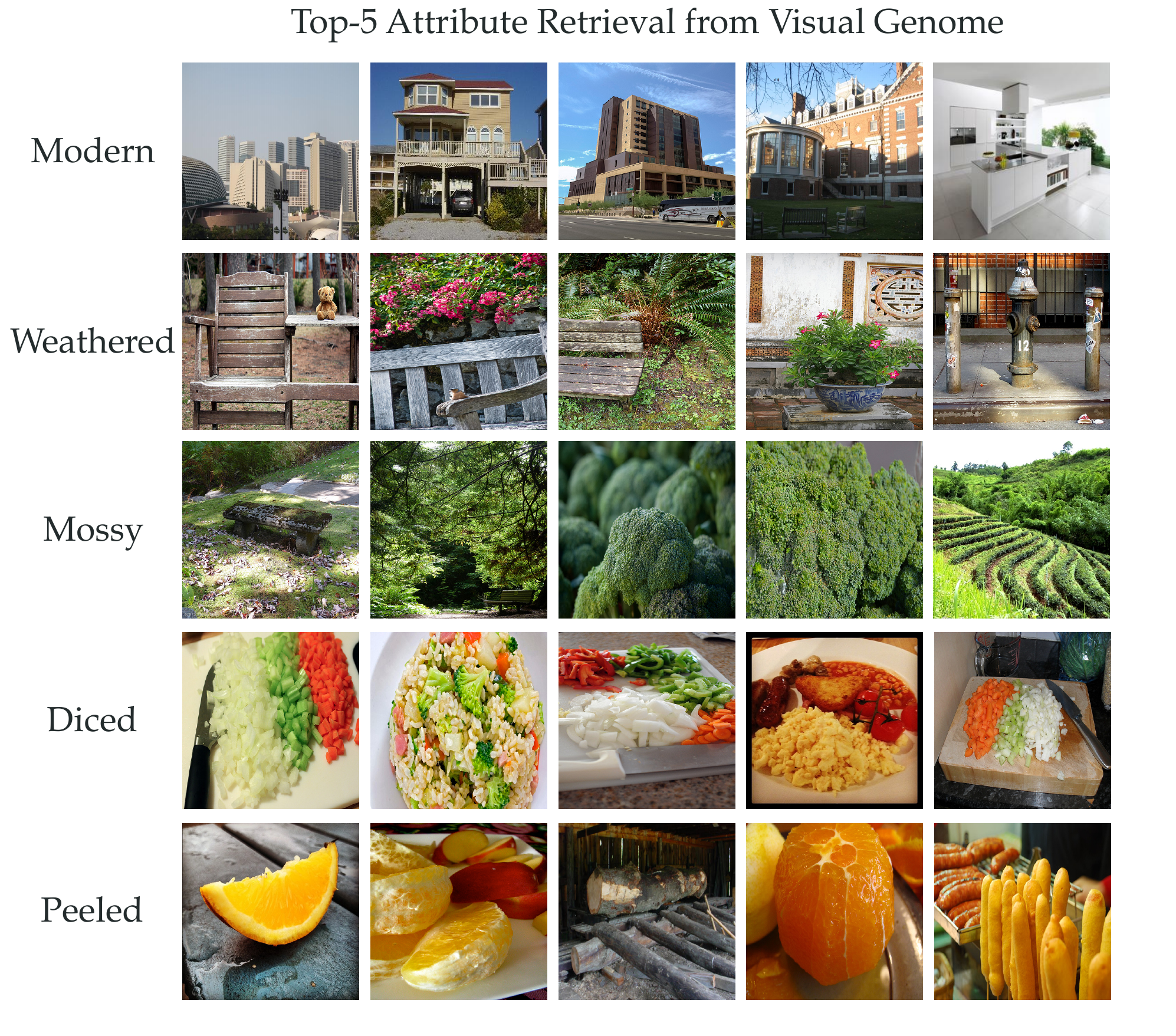}}
            \centerline{(c-1) Visual Genome, attributes}
        \end{minipage}
        \vfill
        \begin{minipage}{.33\linewidth}
            \centerline{\includegraphics[width=\linewidth]{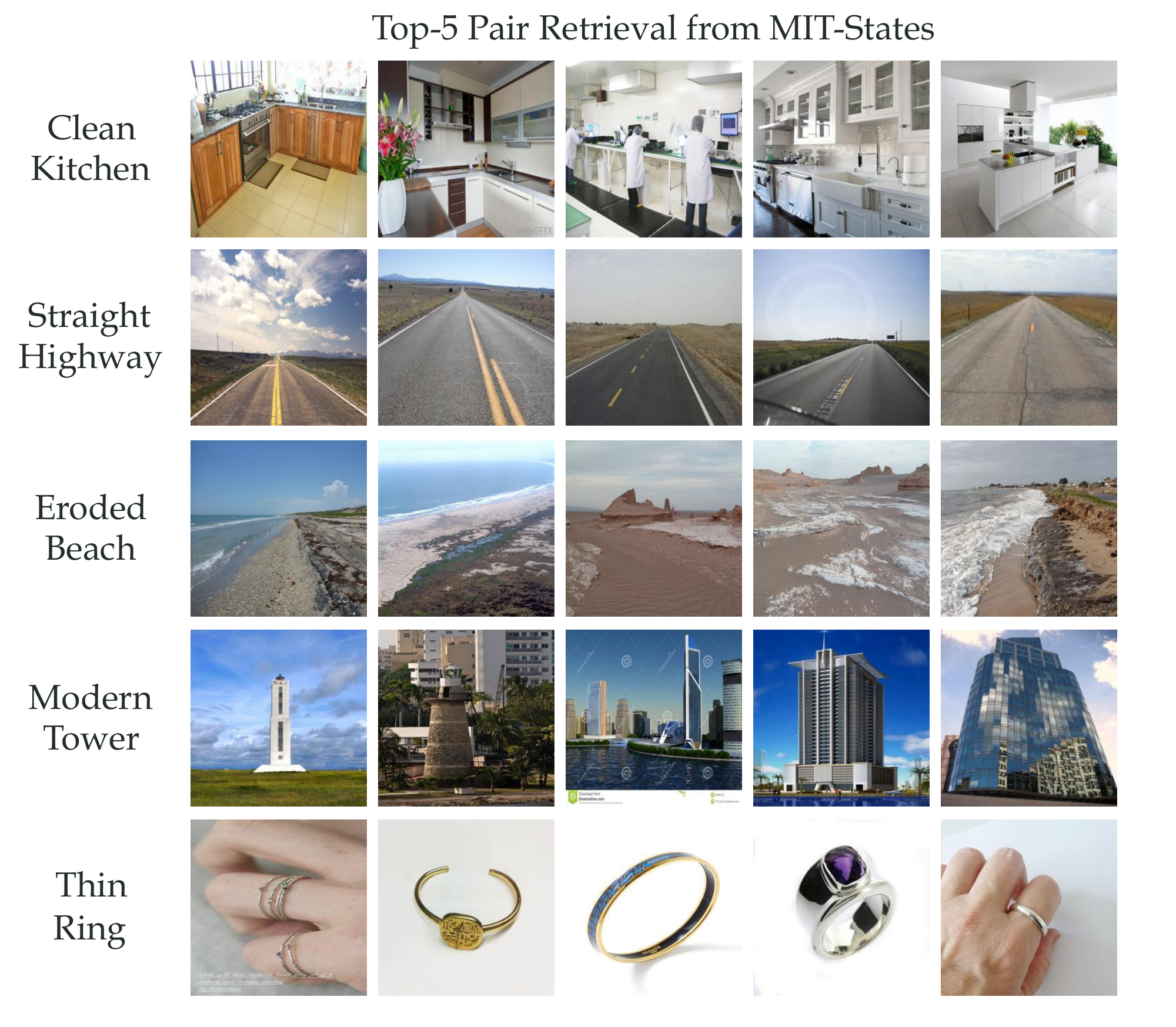}}
            \centerline{(a-2) MIT-States, pairs}
        \end{minipage}
        \hfill
        \begin{minipage}{.33\linewidth}
            \centerline{\includegraphics[width=\linewidth]{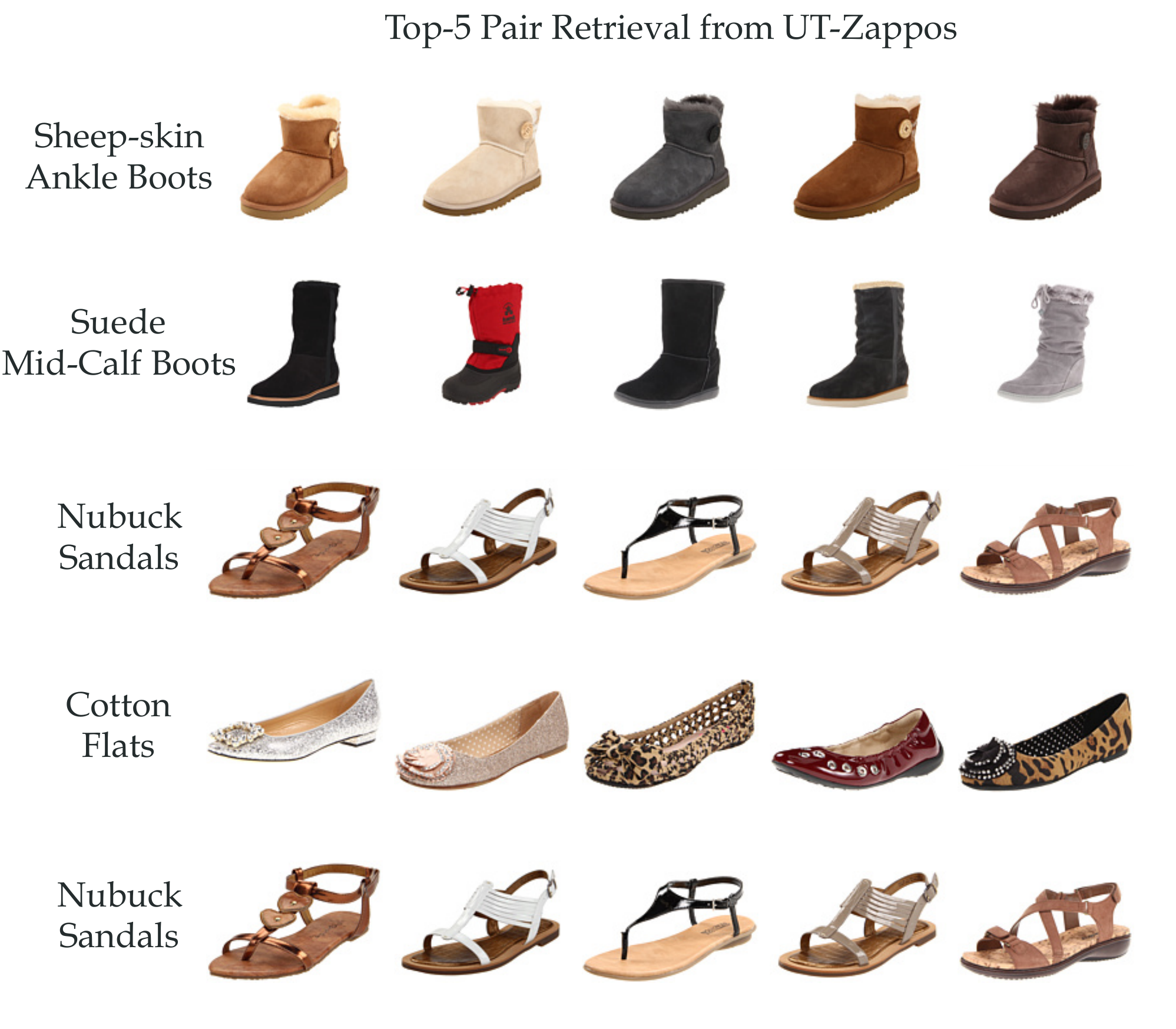}}
            \centerline{(b-2) UT-Zappos, pairs}
        \end{minipage}
        \hfill
        \begin{minipage}{.33\linewidth}
            \centerline{\includegraphics[width=\linewidth]{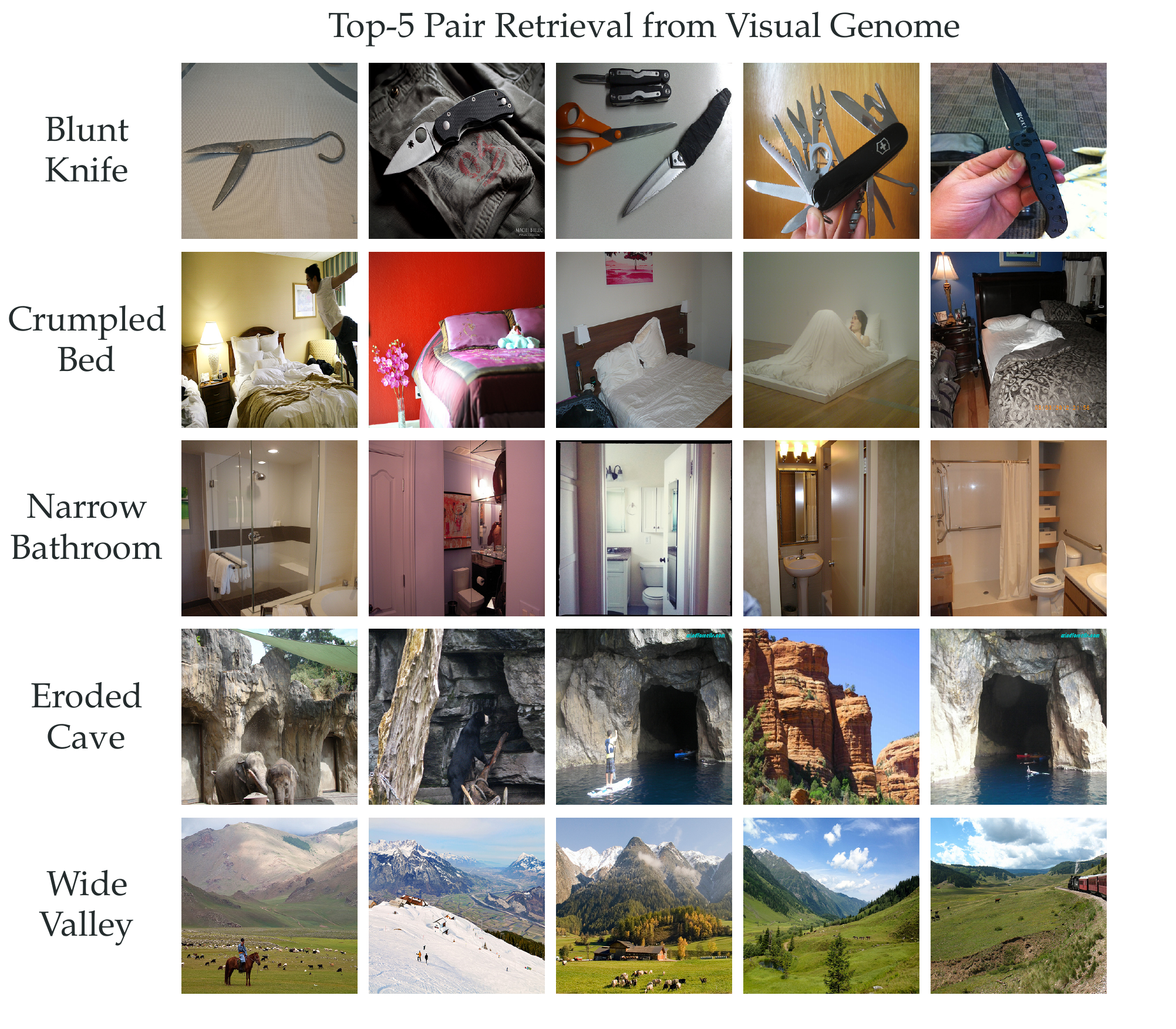}}
            \centerline{(c-2) Visual Genome, pairs}
        \end{minipage}
	\end{center}
	\caption{Additional image retrievals on MIT-States, UT-Zappos (in-domain) and Visual Genome (out-of-domain).}
	\label{Figure:retrieval_add} 
\end{figure*}
In this section, we report normal attribute-object image retrieval results of our method in Fig.~\ref{Figure:retrieval_add}, which contains the in-domain attributes or unseen compositions for UT-Zappos~\cite{ut} and MIT-States~\cite{mit} and out-of-domain retrieval for Visual Genome~\cite{visualgenome}.
We follow the settings of \cite{operator}: 
1) \textbf{In-domain attributes or unseen compositions}: we train SymNet on MIT-States or UT-Zappos and query the attributes or unseen pairs upon the test set of each dataset. 
2) \textbf{Out-of-domain retrieval}: with SymNet \textit{only trained on MIT-States}, we conduct retrieval on the large-scale Visual Genome~\cite{visualgenome} with over 100K images, which is non-overlapping with the training set of MIT-States.

SymNet performs robustly on both in-domain and out-of-domain retrievals.
Our model is capable of recognizing the images with queried attributes and pairs in most cases. When querying an attribute, the model accurately retrieves images across various objects, \eg for MIT-States, the top-5 retrievals of \texttt{fresh} vary among \texttt{fresh-egg}, \texttt{fresh-milk} and \texttt{fresh-flower}. 
In out-of-domain retrieval, our SymNet also shows its robustness. Though it has never seen the images in Visual Genome, the model generalizes well on the target domain and returns correct retrievals, \eg \texttt{dark} objects and \texttt{unripe lemon}.

\begin{figure*}[!ht]
    \begin{center}
        \begin{minipage}{1.\linewidth}
            \centerline{\includegraphics[width=\linewidth]{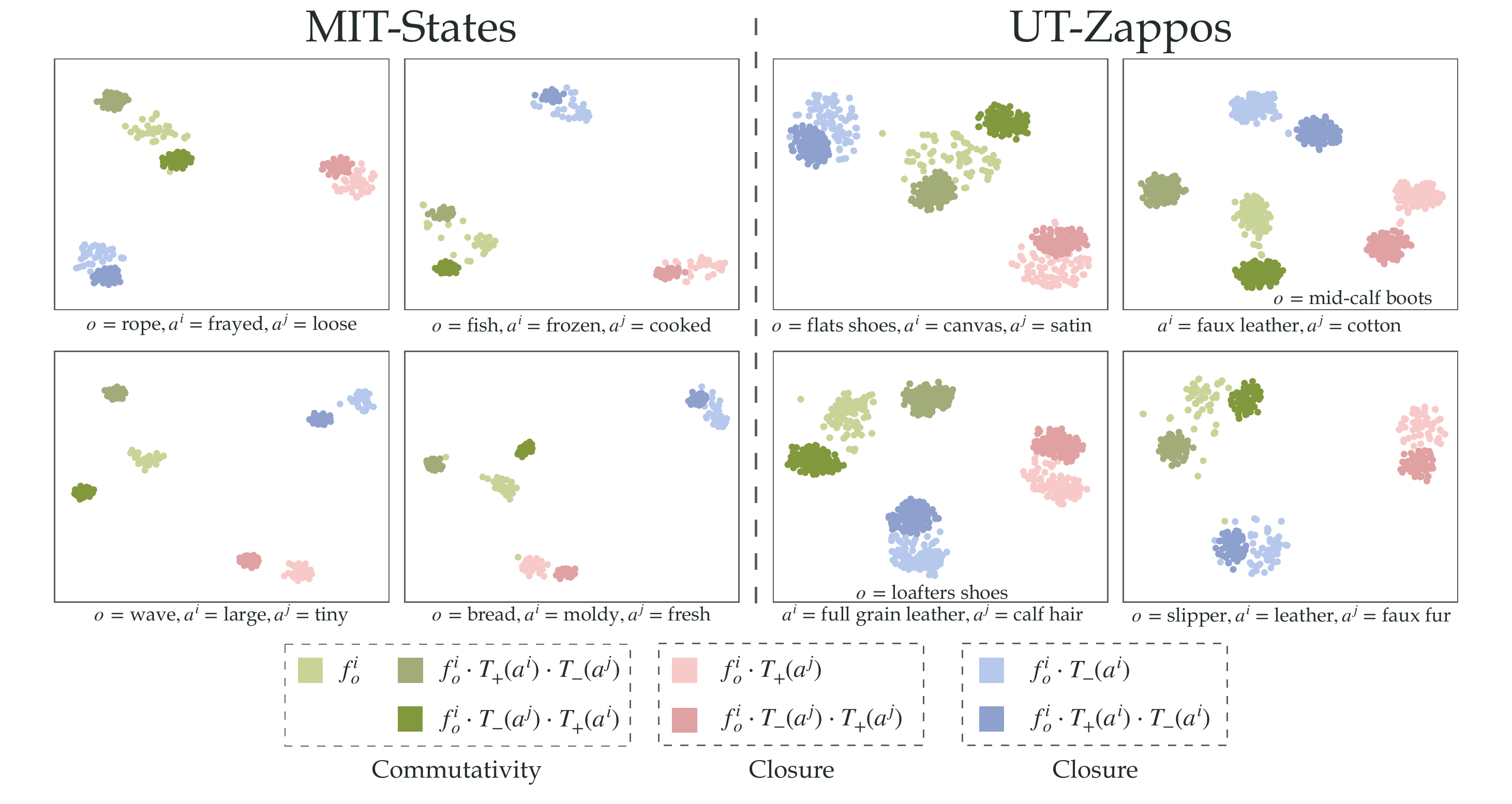}}
            \vspace{-0.15cm}
            \centerline{\small (a) Closure and Commutativity}
        \end{minipage}
        \vfill
        \begin{minipage}{1.\linewidth}
            \centerline{\includegraphics[width=\linewidth]{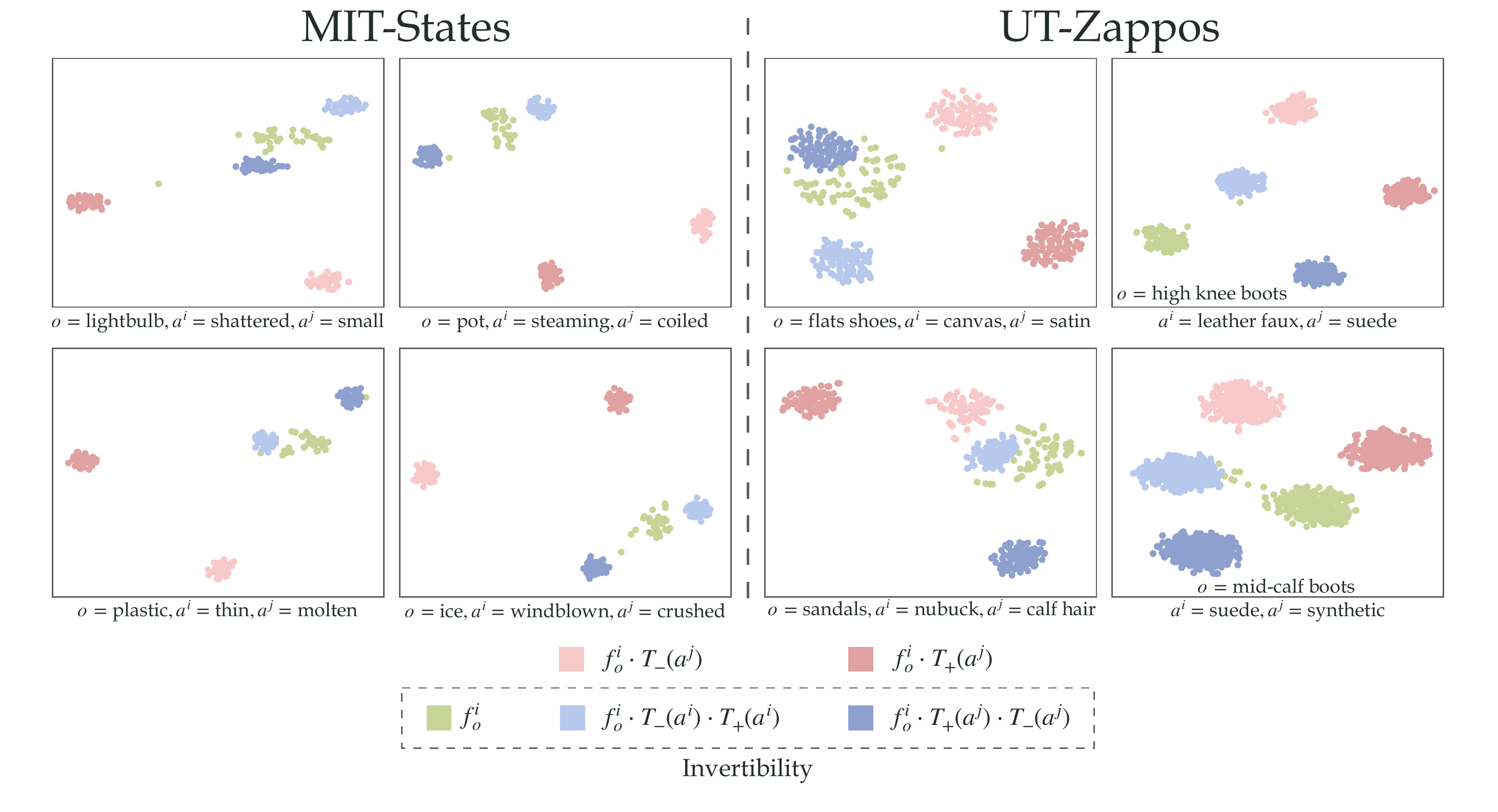}}    \vspace{-0.15cm}
            \centerline{\small (b) Invertibility}
        \end{minipage}
	\end{center}
	\caption{Visualizations of the transformations to verify group axioms. The points with colors in a same dotted box should be close.} 
	\label{Figure:tsne_axiom} 
	\vspace{-0.5cm}
\end{figure*}

\section{Visualized Transformations}
\begin{figure*}[!ht]
    \begin{center}
        \begin{minipage}{1.\linewidth}
            \centerline{\includegraphics[width=\linewidth]{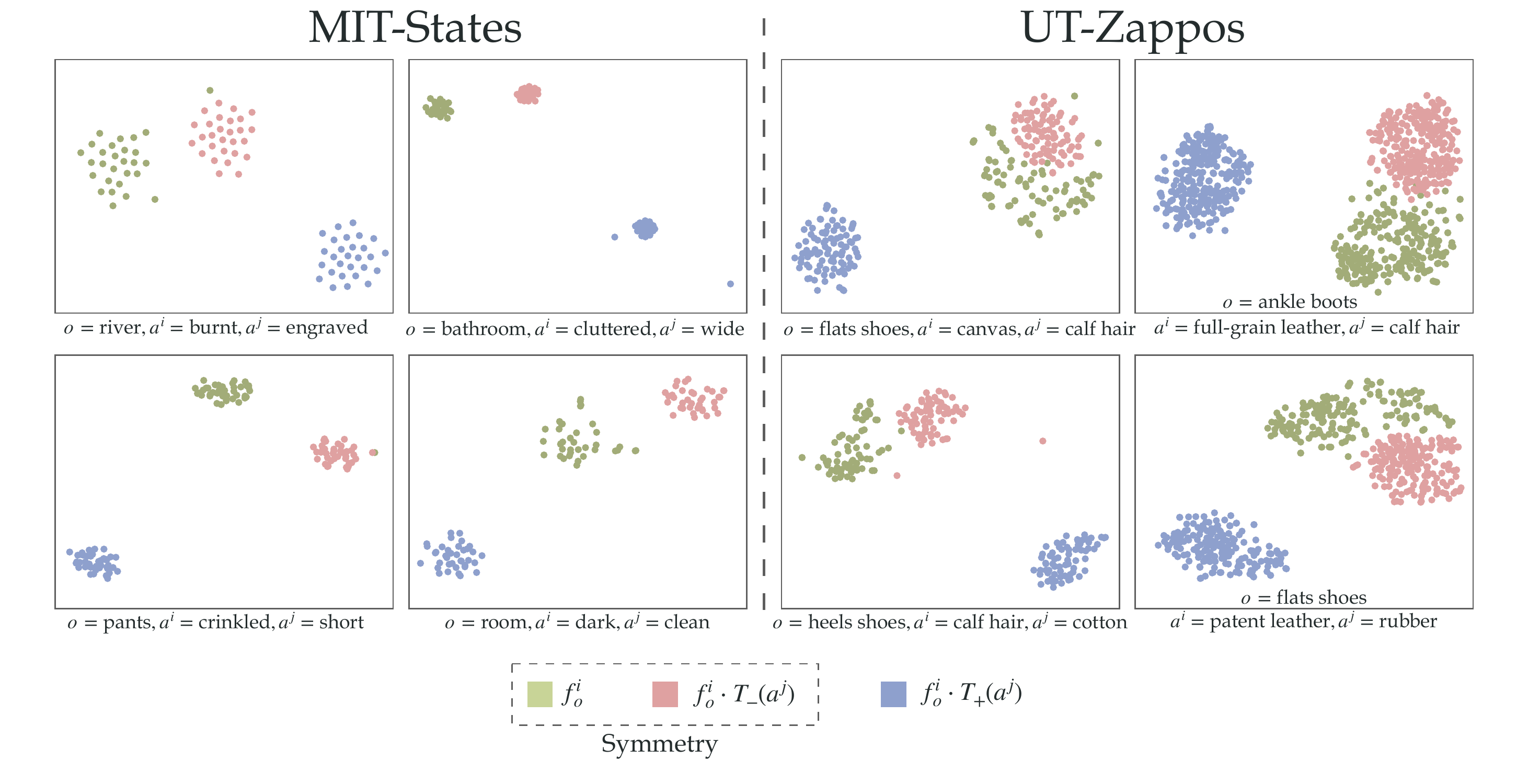}}
            \centerline{\small (a) Symmetry-1}
        \end{minipage}
        \vfill
        \begin{minipage}{1.\linewidth}
            \centerline{\includegraphics[width=\linewidth]{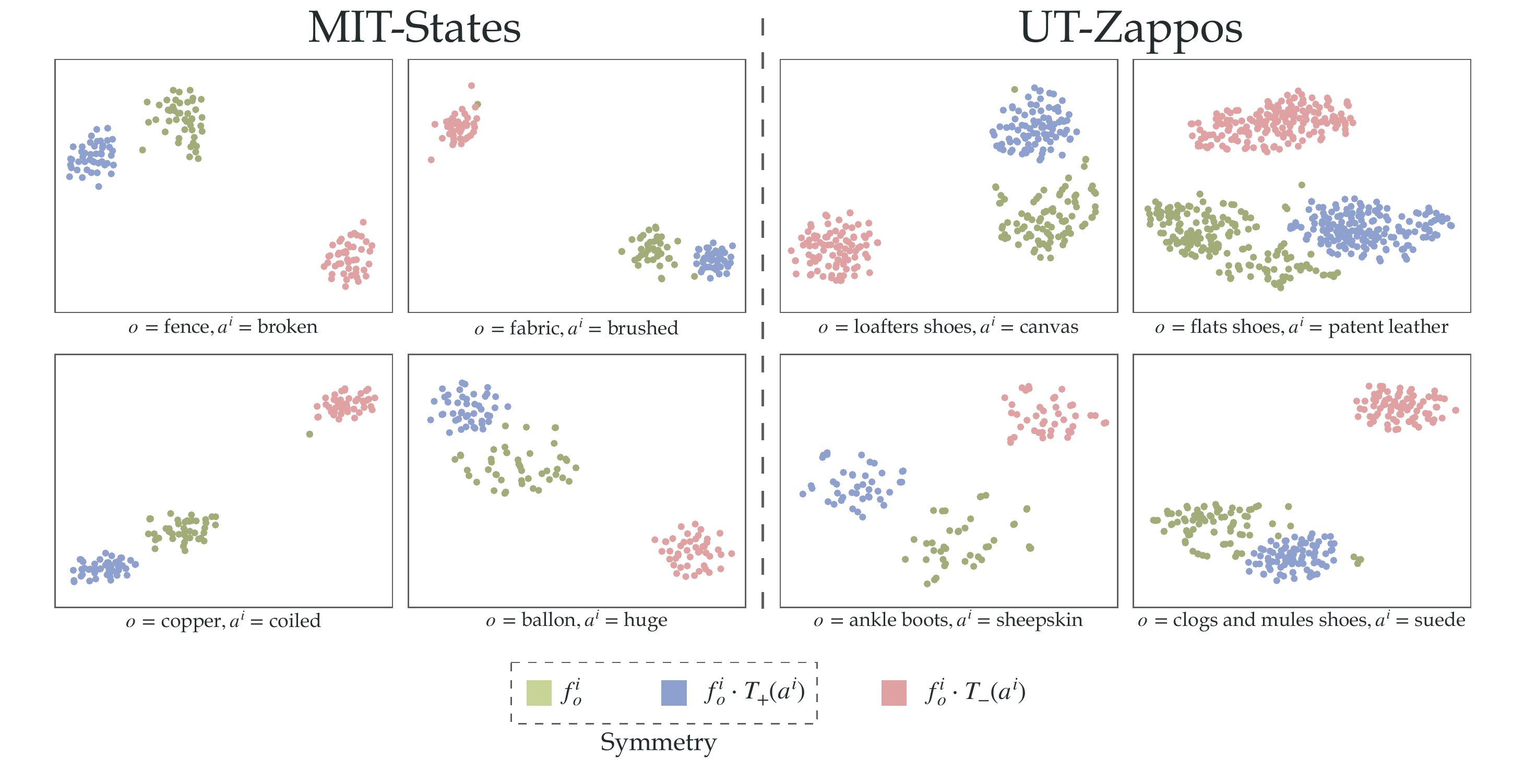}}
            \centerline{\small (b) Symmetry-2}
        \end{minipage}
	\end{center}
	\caption{Visualizations of the transformations to verify symmetry property. The points with colors in a same dotted box should be close.} 
	\label{Figure:tsne_symmetry}
\end{figure*}

In addition, we also provide more visualized transformations of the attribute-object compositions via t-SNE~\cite{tsne} in Fig.~\ref{Figure:tsne_axiom} and Fig.~\ref{Figure:tsne_symmetry}. We observe that the proposed $\{T_e,T_+,T_-\}$ can robustly operate the transformations. The axiom objectives and relative moving distance (RMD) rules are well satisfied during the embedding transformations.

\section{Analysis of Dataset}
Comparatively, accuracy on MIT-States is much lower than UT-Zappos as MIT-States has much more object and attribute categories and suffers from noisy samples and data insufficiency.

Besides, the synonyms and near-synonyms in attributes greatly affect the results. For example, SymNet recognizes $20.4\%$ samples with attribute \texttt{ancient} as \texttt{old}, while the visual properties of these two attributes can barely be distinguished. 
These results are basically correct from the human perspective but mistaken according to the benchmark. 
To explore this phenomenon on MIT-States, we manually select 13 sets of near-synonyms from MIT-States\footnote{\{cracked, shattered, splintered\}; \{chipped, cut\}; \{dirty, grimy\}; \{eroded, weathered\}; \{huge, large\}; \{melted, molten\}; \{ancient, old\}; \{crushed, pureed, mashed\}; \{ripped, torn\}; \{crinkled, crumpled, ruffled, wrinkled\}; \{small, tiny; damp, wet\}
}, which are chosen according to the similarity in both linguistic meanings and visual patterns. We then regard the attributes within each set as equal, \ie, predicting the near-synonym is also considered correct. On this new benchmark, our model achieves $3.03\%$ improvement on attribute accuracy and $0.66\%$ improvement on CZSL accuracy. We also apply this strategy to AttrOperator~\cite{operator}, obtain improvement of $2.25\%$ on attribute recognition and $0.28\%$ on CZSL recognition. Comparing to AttrOperator, our model suffers more from the synonym problem.

\end{appendices}

\end{document}